\documentclass{article}

\newcommand{\EvoPlanRoot}{}
\IfFileExists{EvoPlan-CR/macros.tex}{\renewcommand{\EvoPlanRoot}{EvoPlan-CR/}}{}
\let\EvoPlanBibliographyStyle\bibliographystyle
\renewcommand{\bibliographystyle}[1]{\EvoPlanBibliographyStyle{\EvoPlanRoot#1}}
\usepackage[final]{\EvoPlanRoot corl_2026} %
\let\bibliographystyle\EvoPlanBibliographyStyle
\usepackage{algorithm}
\usepackage{algpseudocode}
\usepackage{placeins}
\usepackage{amsmath,amssymb,mathtools}
\usepackage{booktabs}
\usepackage{multirow}
\usepackage{tabularx}
\usepackage{array}
\usepackage{xspace}
\usepackage{url}
\usepackage{xcolor}
\usepackage{graphicx}
\usepackage{tikz}
\usepackage{enumitem}
\usepackage{listings}
\usepackage{caption}
\usepackage{subcaption}
\usepackage[most]{tcolorbox}
\usetikzlibrary{arrows.meta,positioning,shapes.geometric,fit,calc}

\definecolor{DeepBlue}{HTML}{1F4E79}
\definecolor{DeepGreen}{HTML}{2E6B4E}
\definecolor{DeepPurple}{HTML}{5B2A86}
\definecolor{DeepOrange}{HTML}{B45F06}
\definecolor{LightBlue}{HTML}{EAF3F8}
\definecolor{LightGreen}{HTML}{EAF6EF}
\definecolor{LightPurple}{HTML}{F3EAFE}
\definecolor{LightOrange}{HTML}{FFF3E6}
\definecolor{LineGray}{HTML}{A0A7B0}
\definecolor{CodeBack}{HTML}{F7F7F7}

\setlist[itemize]{leftmargin=1.2em,itemsep=1pt,topsep=2pt,parsep=0pt}
\setlist[enumerate]{leftmargin=1.35em,itemsep=1pt,topsep=2pt,parsep=0pt}

\newcommand{\val}{\textsc{VAL}\xspace}
\newcommand{\llm}{\textsc{LLM}\xspace}

\newcommand{\ourmethod}{EvoPlan\xspace}

\lstdefinestyle{compactcode}{
  basicstyle=\ttfamily\scriptsize,
  backgroundcolor=\color{CodeBack},
  frame=single,
  framerule=0.2pt,
  rulecolor=\color{LineGray},
  columns=fullflexible,
  keepspaces=true,
  breaklines=true,
  xleftmargin=2pt,
  xrightmargin=2pt,
  aboveskip=2pt,
  belowskip=2pt
}

\newcolumntype{Y}{>{\raggedright\arraybackslash}X}

\tikzset{
  module/.style={rectangle, rounded corners, draw=#1, fill=#1!10, thick, align=center, inner sep=3.5pt, minimum height=0.54cm, text width=1.72cm, font=\scriptsize},
  widemodule/.style={rectangle, rounded corners, draw=#1, fill=#1!10, thick, align=center, inner sep=3.5pt, minimum height=0.58cm, text width=2.18cm, font=\scriptsize},
  tinyarrow/.style={-{Latex[length=2mm]}, thick, draw=LineGray},
  dashedgroup/.style={draw=#1, dashed, rounded corners, inner sep=4pt}
}

\title{EvoPlan: Evolutionary Neuro-Symbolic Robot Planning with Spatio-Temporal Guarantees}

\author{
  Bhavya Sai Nukapotula \quad Samin Moosavi \quad Haoze Wang \quad Luke Duncan \\
  \textbf{Diya Shakkottai\thanks{Student at Westwood High School, Austin, TX. Work done as an intern at Texas A\&M University.} \quad Varun Murali \quad Srinivas Shakkottai} \\
  Texas A\&M University \\
  \texttt{\char123 bhavya\_sai, saminmoosavi, haozewang, lukeduncan,} \\ \texttt{mvarun, sshakkot\char125 @tamu.edu}
}

\begin{document}
\maketitle

\begin{abstract}
LLM-based robot planners are fluent but cannot guarantee that their plans are executable or safe.
Classical PDDL planners can guarantee symbolic validity, but only once the problem is fully specified, and they make poor use of an LLM's ability to read context and repair plans.
We present EvoPlan, a neuro-symbolic framework whose guarantees are symbolic validity and satisfaction of a constraint learned from data.
The planner works with locally-hosted open-weight models, so it does not require a cloud service.
EvoPlan has three parts.
First, an \emph{offline procedure that mines a single global Signal Temporal Logic (STL) mobility constraint} from demonstrations.
Perturbed demonstrations supply the missing negatives without encoding any rule, and evolutionary search fits the constraint, which captures driving rules or social-navigation preferences within a modelled taxonomy of norms.
We use the mined constraint to shield driving and navigation policies on Bench2Drive and HA-VLN-CE.
Second, an \emph{evolutionary PDDL planner}: an LLM proposes and repairs plans, programmatic validators decide which ones survive, and the validated portion of the plan grows over iterations.
On open-world ALFWorld and NL-PDDL Blocksworld, it approaches classical planning when the goal vocabulary agrees with the action model, and it stays robust when the two disagree and classical planning fails.
Third, a \emph{constrained execution loop}: Nav2 uses the costmap to ground the plan into waypoints, which are checked against the mined constraint. Violations and execution failures update the PDDL problem for re-planning.
We demonstrate the full pipeline in Gazebo and on a Clearpath Jackal robot.
More details and videos are on the \href{https://nbhavyasai.github.io/evoplan/}{project page}.
\end{abstract}


\section{Introduction}
\label{sec:intro}

LLM and vision-language models can transform natural-language instructions, images, and demonstrations into plausible action proposals, but a plan produced by an LLM alone is not guaranteed to be executable or safe. Classical PDDL planners give machine-checkable guarantees, but only after a problem is fully formalised, and they make little use of an LLM's ability to read context and propose repairs. We couple neural plan generation and repair with symbolic validation and a temporal-logic constraint mined from data to decide which candidates may be executed. The resulting guarantee has two levels. At the task level, every executed plan $\pi$ is valid for the PDDL domain $D$ and problem $P$. At the motion level, every waypoint sequence is checked against the mined STL specification $\Phi_{\mathrm{mob}}$ before execution and monitored online during execution. Safety is thus defined relative to $\Phi_{\mathrm{mob}}$: in the execution loop, a waypoint sequence that violates $\Phi_{\mathrm{mob}}$ is not executed and triggers re-planning, and when $\Phi_{\mathrm{mob}}$ shields an existing policy, an action whose predicted rollout violates the constraint is vetoed and replaced by the candidate action with the highest robustness margin.

\begin{figure}[t]
\centering
\resizebox{0.98\linewidth}{!}{%
\begin{tikzpicture}[x=1cm,y=1cm]
  \path[use as bounding box] (-0.95,-2.35) rectangle (10.35,0.85);
  \tikzset{sysnode/.style={draw,rounded corners,minimum height=0.46cm,text width=1.38cm,align=center,font=\scriptsize,inner sep=2pt}}
  \node[sysnode,fill=DeepBlue!12,draw=DeepBlue!70] (demos) at (0,0) {expert\\demos};
  \node[sysnode,fill=DeepBlue!12,draw=DeepBlue!70] (signals) at (1.85,0) {kinematic +\\social signals};
  \node[sysnode,fill=DeepBlue!12,draw=DeepBlue!70] (neg) at (3.70,0) {perturbed\\negatives};
  \node[sysnode,fill=DeepBlue!12,draw=DeepBlue!70] (sets) at (5.55,0) {contrastive\\sets $\mathcal{D}^+/\mathcal{D}^-$};
  \node[sysnode,fill=DeepPurple!12,draw=DeepPurple!70] (fit) at (7.40,0) {evolutionary\\STL fitting};
  \node[sysnode,fill=DeepPurple!12,draw=DeepPurple!70] (constraint) at (9.25,0) {global mobility\\constraint};

  \node[sysnode,fill=DeepOrange!12,draw=DeepOrange!70] (task) at (9.25,-1.18) {task + PDDL\\domain/state};
  \node[sysnode,fill=DeepOrange!12,draw=DeepOrange!70] (solver) at (7.40,-1.18) {evolutionary\\PDDL solver};
  \node[sysnode,fill=DeepOrange!12,draw=DeepOrange!70] (val) at (5.55,-1.18) {PDDL\\validation};
  \node[sysnode,fill=DeepGreen!12,draw=DeepGreen!70] (exec) at (3.70,-1.18) {trajectory\\planner};
  \node[sysnode,fill=DeepPurple!12,draw=DeepPurple!70] (check) at (1.85,-1.18) {global STL\\check};
  \node[sysnode,fill=DeepGreen!12,draw=DeepGreen!70] (monitor) at (0,-1.18) {monitor +\\replan};

  \draw[tinyarrow] (demos) -- (signals);
  \draw[tinyarrow] (signals) -- (neg);
  \draw[tinyarrow] (neg) -- (sets);
  \draw[tinyarrow] (sets) -- (fit);
  \draw[tinyarrow] (fit) -- (constraint);
  \draw[tinyarrow] (constraint) -- (task);
  \draw[tinyarrow] (task) -- (solver);
  \draw[tinyarrow] (solver) -- (val);
  \draw[tinyarrow] (val) -- (exec);
  \draw[tinyarrow] (exec) -- (check);
  \draw[tinyarrow] (check) -- (monitor);
  \draw[tinyarrow,rounded corners,DeepGreen!80] (monitor.south) -- (0,-2.17)
    -- node[fill=white,inner sep=1.5pt,font=\scriptsize] {failure feedback} (9.25,-2.17) -- (task.south);

  \draw[dashed,rounded corners,DeepBlue!75] (-0.95,-0.48) rectangle (10.20,0.48);
  \node[font=\scriptsize\bfseries\color{DeepBlue}] at (4.60,0.64) {Offline constraint extraction};
  \draw[dashed,rounded corners,DeepOrange!80] (-0.95,-1.72) rectangle (10.20,-0.70);
  \node[font=\scriptsize\bfseries\color{DeepOrange}] at (5.00,-1.86) {Evolutionary PDDL solving and constrained deployment};
\end{tikzpicture}}
\caption{System overview. \emph{Offline}, kinematic and social signals from expert demonstrations and their perturbations form contrastive sets $\mathcal{D}^+/\mathcal{D}^-$. Evolutionary STL fitting produces a global mobility constraint. \emph{Online}, the evolutionary PDDL solver proposes a plan that passes symbolic validation. A trajectory planner expands it into waypoints for the STL check and monitored execution. Motion failures update the PDDL state and trigger replanning.}
\label{fig:system_overview}
\end{figure}

Figure~\ref{fig:system_overview} shows the offline mining pipeline (top row) and the online plan-and-check loop (bottom row). VAL feedback guides symbolic correctness, while motion checks and execution failures inform plan repair. Thus, executability influences the PDDL plan through feedback without encoding continuous motion constraints in the action model. We instantiate this view with three contributions.
\begin{enumerate}[label=\textbf{(\arabic*)}, leftmargin=*, topsep=2pt, itemsep=2pt, parsep=0pt]
\item \textbf{Offline STL constraint mining} (Sec.~\ref{sec:learning}). We mine a single global Signal Temporal Logic (STL) constraint $\Phi_{\mathrm{mob}}$ from demonstrations. All recorded traces are acceptable behaviour, making the data one-class. We synthesise negatives by perturbing demonstrations, procedurally and with an LLM, into statistically different traces. The perturbations specify only a signal's adverse direction, leaving thresholds and rules to be learned from data through supervised classification. The constraint is \emph{global}: its bounds on speed, clearance, and social spacing apply at all times, independently of action schemas, supporting candidate-plan evaluation and constraining every mobility action during execution. Within the taxonomy of norms we model, the same procedure recovers codified \emph{rules} (mined from nuPlan~\citep{caesar2021nuplan} driving logs) or population \emph{preferences} (mined from SCAND~\citep{karnan2022scand} teleoperation), depending on the data.
\item \textbf{Evolutionary PDDL planning with STL feedback} (Sec.~\ref{sec:planning}). The planner uses locally-hosted open-weight models (Qwen3-32B, gpt-oss-120b, and Gemma~4~31B), requiring no cloud service. Rather than calling a classical search procedure, an LLM proposes and repairs plans against a fixed PDDL problem $P$ and domain $D$, and a programmatic validator cascade scores each candidate. For robot tasks, VAL checks symbolic correctness and STL checks assess the costmap-grounded candidate motions. Both provide feedback for evolutionary selection and repair before execution. Elitism on validator feedback grows the validated plan prefix over iterations without explicit suffix-only mutation. When the goal vocabulary agrees with the action model, a classical planner matches or beats ours: Fast Downward~\citep{helmert2006fastdownward} solves every aligned ALFWorld task and all non-misaligned Blocksworld variants. Our planner's advantage appears when the two vocabularies disagree: the LLM bridges them, and the planner solves $98$--$100\%$ of tasks on misaligned ALFWorld and Blocksworld Misalignment, where Fast Downward solves $44\%$ and $0\%$. On Mystery Blocksworld, validate-and-repair solves more instances, but population search reduces excess actions over the optimum by $44\%$.
\item \textbf{Constrained execution loop} (Sec.~\ref{sec:runtime_bridge}). Nav2 grounds each action $a_k$ of the selected plan $\pi=(a_1,\ldots,a_n)$ in the current costmap. Its waypoint sequence $w_k$ is rechecked against $\Phi_{\mathrm{mob}}$ before execution. A violation rejects $a_k$, commits the verified prefix, updates the initial state, and re-invokes the planner. During execution, an online monitor emits failure facts that update $P$ and trigger the same re-planning. The loop runs in Gazebo and on a Clearpath Jackal robot.
\end{enumerate}

The mined constraints shield a Qwen2.5-VL driver on Bench2Drive and two discrete-action navigation policies on HA-VLN-CE (Sec.~\ref{sec:downstream_eval}). The planner is compared with LLM-only planners, NL-PDDL, and Fast Downward on open-world and closed-world ALFWorld and the NL-PDDL~\citep{liu2026nlpddl} Blocksworld variants (Sec.~\ref{sec:exp_planner}). The full system runs in Gazebo and on a Jackal robot (Sec.~\ref{sec:exp_endtoend}).

\section{Related Work}
\label{sec:related}

\paragraph{LLM/VLM planning with formal representations.}
Recent work combines foundation models with formal planners to overcome the brittleness of one-shot natural-language planning. VLMFP generates PDDL domain and problem files from visual observations using a dual-VLM loop, then checks symbolic execution against simulated outcomes~\citep{hao2026simulation}. Code-as-Symbolic-Planner prompts an LLM to write code that serves as its own symbolic solver and constraint checker~\citep{chen2025codeassymbolic}. VisualPredicator learns neuro-symbolic predicates from interaction to build abstract world models for symbolic task planning~\citep{liang2025visualpredicator}. NL-PDDL replaces rigid symbolic predicates with natural-language counterparts and uses lifted regression with commonsense entailment to handle open-world goal-action misalignment~\citep{liu2026nlpddl}. PIP-LLM uses PDDL for team-level task decomposition and integer programming for robot-level allocation in multi-robot coordination~\citep{shi2025pipllm}. These systems use formal intermediate languages. However, they check plans symbolically, in simulation, or with LLM-authored checkers. None checks the executed continuous trace against a constraint mined from demonstrations.

\paragraph{Safety-aware symbolic planning and verification.}
Classical PDDL planning separates domain schemas from problem instances~\citep{mcdermott1998pddl}, and validators such as VAL provide independent checks of symbolic plan validity~\citep{howey2004val}. SafeGen-LLM~\citep{fan2026safegen} trains LLMs with verifier-derived rewards to satisfy PDDL3 state-trajectory constraints~\citep{gerevini2009pddl3}, which express safety over the discrete states a plan visits. In contrast, we mine an STL constraint from demonstrations and enforce it on each action's waypoint trace before and during execution, constraining execution rather than certifying symbolic effects.

\paragraph{Temporal logic learning, evolutionary search, and robot safety.}
STL provides a language for temporal properties over real-valued signals~\citep{maler2004monitoring}, and robustness semantics support quantitative monitoring~\citep{donze2010robust}. Recent methods learn temporal-logic predicates from data with statistical guarantees~\citep{soroka2025learning}, motivating our constraint-identification stage. TeLEx learns STL parameters from positive examples alone using a tightness metric~\citep{jha2019telex}. Our demonstrations are also one-class, but we add perturbed demonstrations as negatives so that the constraint is fit discriminatively. AlphaEvolve demonstrates that LLM-guided evolutionary search can improve machine-gradable symbolic artifacts through automatic evaluation~\citep{novikov2025alphaevolve}. We apply this pattern to PDDL plans and STL constraints. Complementary work on grounded heterogeneous teams~\citep{ravichandran2025spineht} and robot constitutions~\citep{sermanet2025robotconstitutions} highlights the need for closed-loop grounding and semantic safety in real robot systems. Further comparisons appear in Appendix~\ref{app:related_extended}.

\section{Problem Definition}
\label{sec:problem}

We address the problem of executing a natural-language task in a partially known, dynamic world while satisfying \emph{implicit norms} that are not stated in the task itself, given only \emph{unannotated} prior robot operation logs as the source of those norms.

\paragraph{Inputs.}
A deployment instance is the tuple $\mathcal{I} = (\mathcal{W},\, \mathcal{A},\, g_{\mathrm{lang}},\, \mathcal{N},\, \mathcal{D})$. The \emph{partially known world model} $\mathcal{W} = (\mathcal{S}, \widehat{T})$ has a state space $\mathcal{S}$ observed only locally through a perception stack and an incomplete transition model $\widehat{T}$: regions are unmapped, semantic labels are noisy, and other-agent dynamics are not given in closed form. A set of \emph{dynamic agents} $\mathcal{A}$ (pedestrians, vehicles, or other moving entities) is observed online but not provided in advance. The \emph{explicit language task} $g_{\mathrm{lang}}$ is expressed in natural language (e.g., ``drive to the cafeteria''). The \emph{implicit norms} $\mathcal{N}$, which the robot must additionally satisfy, are codified rules (e.g., stop at red lights) or population preferences (e.g., maintain social distance), depending on the domain. Crucially, $\mathcal{N}$ is not given in symbolic form, only \emph{implicitly in the data}. The corpus $\mathcal{D} = \{\tau_1, \ldots, \tau_n\}$ of \emph{unannotated} prior operation logs contains time series of signals over robot, environment, and other-agent state. By construction every trace is acceptable behaviour, but no $\mathcal{N}$ annotation is attached to any signal.

\paragraph{Output.}
A \emph{constrained execution trace} that achieves \(g_{\mathrm{lang}}\): a PDDL plan \(\pi=(a_1,\ldots,a_n)\) produced by a discrete planner from a domain \(D\) and a problem \(P\) that encodes \(g_{\mathrm{lang}}\), such that
(i) \(\pi\) is symbolically valid (preconditions, effects, goal reachability),
(ii) each action \(a_k\) is mapped by a navigation layer to a waypoint sequence \(w_k\), and
(iii) the concatenated trace \(w_1\cdots w_n\) satisfies a signal-level constraint that approximates \(\mathcal{N}\). Because \(\mathcal{N}\) is only implicit, this approximation is what can be checked.
When \(w_k\) violates the constraint at runtime, the verified prefix \(a_1\ldots a_{k-1}\) is committed, the resulting world state becomes the new initial state, and the planner is re-invoked to produce a fresh suffix-replacing plan.

\paragraph{What makes this hard.}
Three properties of the setting jointly rule out the standard fixes.
\emph{(1) Partial observability and dynamics.}
The world model is incomplete and the moving agents' trajectories must be tracked online from sensor data, so a planner that requires a fully specified environment cannot be applied directly.
\emph{(2) Underspecification of the prompt.}
The language task \(g_{\mathrm{lang}}\) underspecifies the full normative requirement: a planner that optimises only for \(g_{\mathrm{lang}}\) can satisfy the prompt (``you reached the goal'') while violating \(\mathcal{N}\) (``you ran a red light'', ``you brushed past a pedestrian'').
\emph{(3) One-class data.}
The only prior signal is the unannotated corpus \(\mathcal{D}\) in which every recorded trajectory is acceptable behaviour.
There are no labelled violations, no preference pairs, and no symbolic specification of \(\mathcal{N}\).
Conventional fixes (reward shaping with hand-coded penalties, behaviour cloning, or RL from preference pairs) each require either a manually written specification of \(\mathcal{N}\), explicitly annotated violation data, or pairwise preference labels.
We assume none of these are available.

\paragraph{Approach.}
\emph{Offline}, we approximate \(\mathcal{N}\), within the taxonomy of norms we model, by a global mobility constraint \(\Phi_{\mathrm{mob}}\). Perturbed demonstrations supply the negatives that turn the one-class data into a supervised discrimination problem, and \(\Phi_{\mathrm{mob}}\) is fit over signals to separate demonstrations from their perturbations (Sec.~\ref{sec:learning}).
\(\Phi_{\mathrm{mob}}\) is a single STL formula that applies uniformly to every mobility action. The same procedure produces rule-like \(\Phi_{\mathrm{mob}}\) when \(\mathcal{D}\) encodes a rule-governed domain and preference-like \(\Phi_{\mathrm{mob}}\) when \(\mathcal{D}\) records how a population of operators prefers to move (Sec.~\ref{sec:downstream_eval}).
\emph{Online}, the language task \(g_{\mathrm{lang}}\) is translated into a PDDL problem, and an LLM-driven evolutionary PDDL planner (Sec.~\ref{sec:planning}) uses VAL and, for robot tasks, STL feedback to propose and repair plans until VAL accepts one as valid for the domain.
The navigation layer maps each action \(a_k\) in \(\pi\) to waypoints \(w_k\), which are rechecked against \(\Phi_{\mathrm{mob}}\) before execution.
If \(w_k\) satisfies \(\Phi_{\mathrm{mob}}\), \(a_k\) is scheduled for execution. Otherwise, the verified prefix \(a_1\ldots a_{k-1}\) is committed, the initial state is updated from the resulting world, and the planner is re-invoked with the same goal under the updated state, yielding a new plan whose remaining actions are themselves re-checked under \(\Phi_{\mathrm{mob}}\).
During execution, an online monitor evaluates \(\Phi_{\mathrm{mob}}\) on the measured trace and emits failure facts that trigger the same re-planning.
Guarantees are therefore relative to \(\Phi_{\mathrm{mob}}\). When used as a shield, it vetoes a policy action whose predicted rollout violates the constraint and substitutes the candidate action with the highest robustness margin. The execution loop admits only waypoint sequences that satisfy \(\Phi_{\mathrm{mob}}\) and re-plans otherwise.

\section{EvoPlan Framework}
\label{sec:method}
EvoPlan has three parts (Fig.~\ref{fig:system_overview}): offline mining of the mobility constraint $\Phi_{\mathrm{mob}}$ from demonstrations (Sec.~\ref{sec:learning}), online evolution of a VAL-valid PDDL plan $\pi$ (Sec.~\ref{sec:planning}), and an execution loop that runs $\pi$ only through waypoints that satisfy $\Phi_{\mathrm{mob}}$ and feeds failures back (Sec.~\ref{sec:runtime_bridge}).

\subsection{Offline STL constraint mining}
\label{sec:learning}

We mine a single \emph{global} constraint $\Phi_{\mathrm{mob}}$: a Signal Temporal Logic (STL) formula over named signals such as speed, clearance, and time-to-collision. It constrains every motion of every plan independently of the action schema.

\paragraph{Perturbed negatives.}
Demonstrations are one-class data: every recorded trace is acceptable, so there are no bad examples to fit against. We create them by perturbing each demonstration into a statistically different trace, procedurally (e.g., \texttt{speeding}, \texttt{tailgating}, \texttt{crowding\_front}) and with an \llm\ generator. A perturbation fixes only which direction of a signal is worse, with no threshold or rule, so the demonstrations $\mathcal{D}^+$ and their perturbations $\mathcal{D}^-$ turn mining into a classification problem whose constants come from the data (perturbation modes: Appendix~\ref{app:mining_perturbations}).

\paragraph{Evolutionary fitting.}
An LLM-driven evolutionary search over program edits~\citep{novikov2025alphaevolve} proposes formulas from an STL grammar over these signals (Appendix~\ref{app:mining_grammar}) and scores their robustness $\rho$:
\begin{equation*}
\max_{\Phi\in\mathcal{G}}\; \mathbb{E}_{\tau\in\mathcal{D}^+}\rho(\tau,\Phi)
-\lambda_-\,\mathbb{E}_{\tau\in\mathcal{D}^-}\max(0,\rho(\tau,\Phi))
-\lambda_c\,\mathrm{complexity}(\Phi),
\end{equation*}
rewarding satisfaction on demonstrations, penalising satisfaction on perturbed traces, and penalising complexity. The same procedure recovers codified \emph{rules} from driving logs (nuPlan~\citep{caesar2021nuplan}) and population \emph{preferences} from teleoperation (SCAND~\citep{karnan2022scand}). Only the data and perturbation modes change.

\paragraph{Enforcement.}
Because $\Phi_{\mathrm{mob}}$ is defined over signals rather than inside a controller, it can be enforced without retraining a policy. It serves as a shield that vetoes a policy's action when its short-horizon rollout violates $\Phi_{\mathrm{mob}}$ and substitutes the candidate action whose rollout has the highest robustness (Sec.~\ref{sec:downstream_eval}, Appendix~\ref{app:mining_prefs_shield}), and as the motion check of the execution loop (Sec.~\ref{sec:runtime_bridge}).
\subsection{Evolutionary PDDL planning}
\label{sec:planning}

\begin{figure}[!ht]
\centering
\begin{subfigure}{0.32\linewidth}
  \centering
  \includegraphics[width=\linewidth]{\EvoPlanRoot 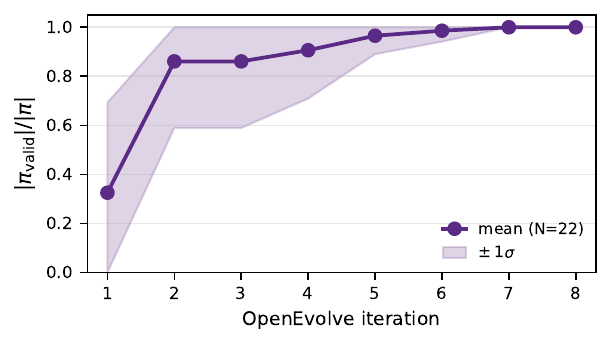}
  \caption{Open-world}
  \label{fig:vp_growth_open}
\end{subfigure}\hfill
\begin{subfigure}{0.32\linewidth}
  \centering
  \includegraphics[width=\linewidth]{\EvoPlanRoot 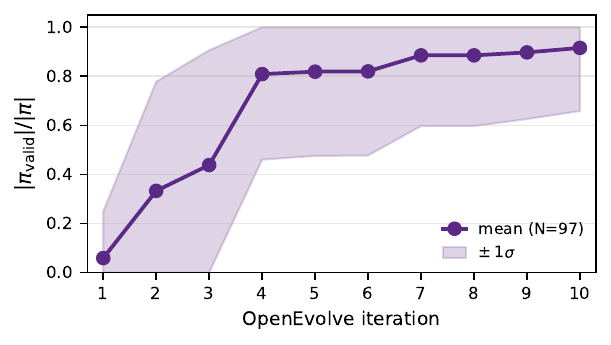}
  \caption{Closed-world}
  \label{fig:vp_growth_closed}
\end{subfigure}\hfill
\begin{subfigure}{0.32\linewidth}
  \centering
  \includegraphics[width=\linewidth]{\EvoPlanRoot 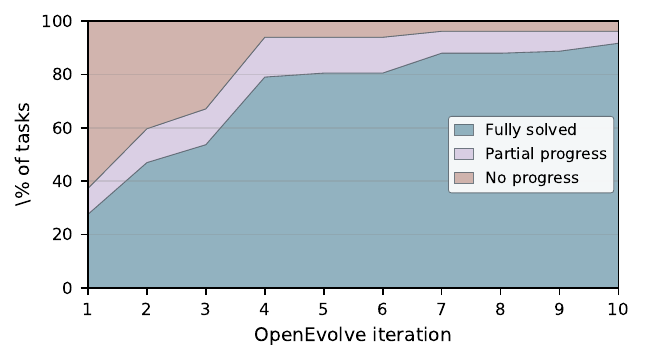}
  \caption{Closed-world population}
  \label{fig:vp_stacked}
\end{subfigure}
\caption{Verified-prefix growth on hard tasks (Qwen3): (a) open-world, (b) closed-world, (c) closed-world population state.}
\label{fig:vp_growth}
\end{figure}

Given a domain $D$ and problem $P$, we evolve a grounded plan $\pi$ with machine-gradable evolution~\cite{novikov2025alphaevolve}: an LLM proposes plans and edits, and a validator decides which survive.  Each generation samples a parent from a population of candidates (islands with elite sampling), and the LLM proposes an edit conditioned on the parent's \emph{validator feedback}, the failing action and its unsatisfied precondition. A cascade scores the child: PDDL syntax, VAL~\cite{howey2004val} simulation, and goal reachability, short-circuiting at the first failure. The symbolic stages give valid plans the highest score, and invalid ones are ranked by the fraction of their actions that execute before the first failure (Appendix~\ref{app:planner_algorithm}). No prefix is frozen, yet this validated fraction grows monotonically over iterations on hard tasks in both open- and closed-world runs (Fig.~\ref{fig:vp_growth}). For robot planning, a fourth stage evaluates the STL robustness of candidate waypoints grounded by Nav2 in the current costmap. VAL and STL feedback jointly guide evolutionary selection and repair before execution (Appendix~\ref{app:loop_evolve_pseudocode}). Thus, motion feasibility shapes the PDDL plan without encoding continuous constraints in its action model. The benchmarks in Sec.~\ref{sec:exp_planner} use only the symbolic stages.

\paragraph{Open-world and misaligned problems.}
In the open world, $P$ encodes the robot's belief over observed objects. A validator failure on an undeclared object triggers exploration, the belief grows, $P$ is rebuilt, and the search resumes. When the goal vocabulary differs from the action vocabulary (e.g., goal \emph{heated} vs.\ action \texttt{toast}), a single LLM call produces a predicate map $\mu$ that rewrites the goal predicates of $P$ before VAL. The same $\mu$ applies to objects observed later (Appendix~\ref{app:planner_algorithm}).
\subsection{Constrained execution}
\label{sec:runtime_bridge}

The execution loop feeds motion feasibility back into symbolic planning. Nav2 uses the current costmap to expand each action $a_k$ into a waypoint sequence $w_k$, which is checked against $\Phi_{\mathrm{mob}}$ before execution. If $w_k$ satisfies $\Phi_{\mathrm{mob}}$ it is dispatched. Otherwise, the verified prefix $a_1\ldots a_{k-1}$ is committed and the planner re-plans from the updated state. During execution, an online monitor evaluates robustness on the measured trace and responds at two time scales. A reactive tier handles signal-level hazards without the planner, inflating the local costmap around a person and re-issuing the navigation goal. A deliberative tier emits a failure fact (e.g., \texttt{route-blocked}), retracts the failed transition from $P$, and re-plans. A VAL-accepted plan is swapped in without restarting the executor.

The same loop handles goals over objects whose location is unknown. Such a problem is unsolvable as written, so VAL truncates the plan to its executable prefix, an exploration survey, which the robot drives while a detector searches. A detection grounds the object in $P$ and triggers a re-plan from the robot's current region. The loop runs as a ROS\,2 adapter on a Jackal robot, in Gazebo and on hardware (implementation: Appendix~\ref{app:loop}).

\section{Experiments}
\label{sec:experiments}
\subsection{Mined STL constraints}
\label{sec:downstream_eval}

We test whether one mining procedure recovers interpretable constraints, codified rules or population preferences depending on the data, and whether, used as a zero-shot shield, these constraints reduce the safety violations of policies trained without them while preserving task performance.

\paragraph{Mined constraints.}
From nuPlan~\citep{caesar2021nuplan} driving logs, mining recovers a small set of rules with right/wrong semantics: a stop-on-red implication, vehicle-clearance and time-to-collision floors, and lane-aware speed caps. From SCAND~\citep{karnan2022scand} sidewalk teleoperation, it recovers a conjunction of $17$ always-clauses that encode social-navigation preferences, e.g.\ $\Box(\texttt{min\_clearance}\ge0.48\,\mathrm{m})$, $\Box(\texttt{ped\_ttc}\ge0.42\,\mathrm{s})$, and $\Box(\texttt{crowd\_count}\ge3 \lor \texttt{min\_clearance}\ge0.6\,\mathrm{m})$. These thresholds are fitted from data, not hand-set (full formula: Appendix~\ref{app:mining_prefs_formula}). The perturbed negatives determine which clauses bind but not their thresholds: re-mining across a $16\times$ range of perturbation magnitudes moves the thresholds by a median of $0.6\%$, while without negatives the weakest violation class is detected at $0.000$ instead of $0.457$ (Table~\ref{tab:negatives_main}, Appendix~\ref{app:mining_negatives}).

\paragraph{Shielding downstream policies.}
We shield three policies the constraint was not mined from: a Qwen2.5-VL~\citep{bai2025qwen25vl} driving stack on Bench2Drive~\citep{jia2024bench2drive} with the nuPlan rules, and a pretrained PointNav (DD-PPO~\citep{wijmans2019ddppo}) and a behaviour-cloned MLP on HA-VLN-CE~\citep{wu2024havln} with the SCAND preferences. The shield predicts a $K{=}5$-step rollout for each candidate action. When the policy's action violates $\Phi_{\mathrm{mob}}$ and a pedestrian or vehicle is close, it vetoes that action and executes the candidate with the highest rollout robustness. A policy-prior weight $\alpha$ trades this off against the policy's own preference, and the proximity threshold is set per policy (Appendix~\ref{app:mining_prefs_shield}).

\begin{table}[!ht]
\centering
\caption{Mined constraints as a zero-shot shield. \textbf{(a)} nuPlan rules on Bench2Drive: DS\,$=$\,Driving Score, Coll\,$=$\,collisions, Infr\,$=$\,total infractions. \textbf{(b)} SCAND preferences on HA-VLN-CE \texttt{val\_unseen}, $N{=}1839$ paired episodes per row: SR\,$=$\,success rate, CT\,$=$\,collision terminations. \textbf{(c)} SCAND mining with and without perturbed negatives: Worst\,$=$\,rejection of the weakest perturbation class. Recall\,/\,Spec.\ are measured on real HA-VLN-CE contacts and contact-free windows (Appendix~\ref{app:mining_negatives}). Best value per column in bold.}
\label{tab:downstream}
\scriptsize
\setlength{\tabcolsep}{3pt}
\begin{subtable}[t]{0.30\linewidth}
\centering
\subcaption{Bench2Drive (rules)}
\label{tab:downstream_b2d}
\begin{tabular}{@{}lccc@{}}
\toprule
Shield & DS\,$\uparrow$ & Coll\,$\downarrow$ & Infr\,$\downarrow$ \\
\midrule
off                 & 62.96          & 14         & 20          \\
$\alpha{=}0$        & 65.61          & \textbf{4} & 7           \\
$\alpha{=}0.05$     & 65.10          & 5          & 9           \\
$\alpha{=}0.30$     & \textbf{66.20} & \textbf{4} & \textbf{6}  \\
\bottomrule
\end{tabular}
\end{subtable}\hfill
\begin{subtable}[t]{0.32\linewidth}
\centering
\subcaption{HA-VLN-CE (preferences)}
\label{tab:downstream_headline}
\begin{tabular}{@{}llcc@{}}
\toprule
Policy & Shield & SR\%\,$\uparrow$ & CT\%\,$\downarrow$ \\
\midrule
PointNav                    & off & \textbf{58.7} & 38.1          \\
\quad\emph{$\alpha{=}0.05$} & on  & 57.6          & \textbf{18.5} \\
MLP                         & off & 32.1          & 30.1          \\
\quad\emph{$\alpha{=}0$}    & on  & \textbf{32.7} & \textbf{5.2}  \\
\bottomrule
\end{tabular}
\end{subtable}\hfill
\begin{subtable}[t]{0.33\linewidth}
\centering
\subcaption{Perturbed negatives (SCAND)}
\label{tab:negatives_main}
\begin{tabular}{@{}lccc@{}}
\toprule
Fitted from & Worst & Recall & Spec. \\
\midrule
Percentile (one-class) & n/a            & 1.000 & 0.428 \\
TeLEx~\citep{jha2019telex} & 0.026      & 0.499 & 0.788 \\
Ours (+ negatives)     & \textbf{0.457} & 0.518 & 0.777 \\
Ours, no negatives     & 0.000          & 0.477 & 0.790 \\
\bottomrule
\end{tabular}
\end{subtable}
\end{table}

\begin{figure}[!ht]
\centering
\includegraphics[width=0.98\linewidth]{\EvoPlanRoot 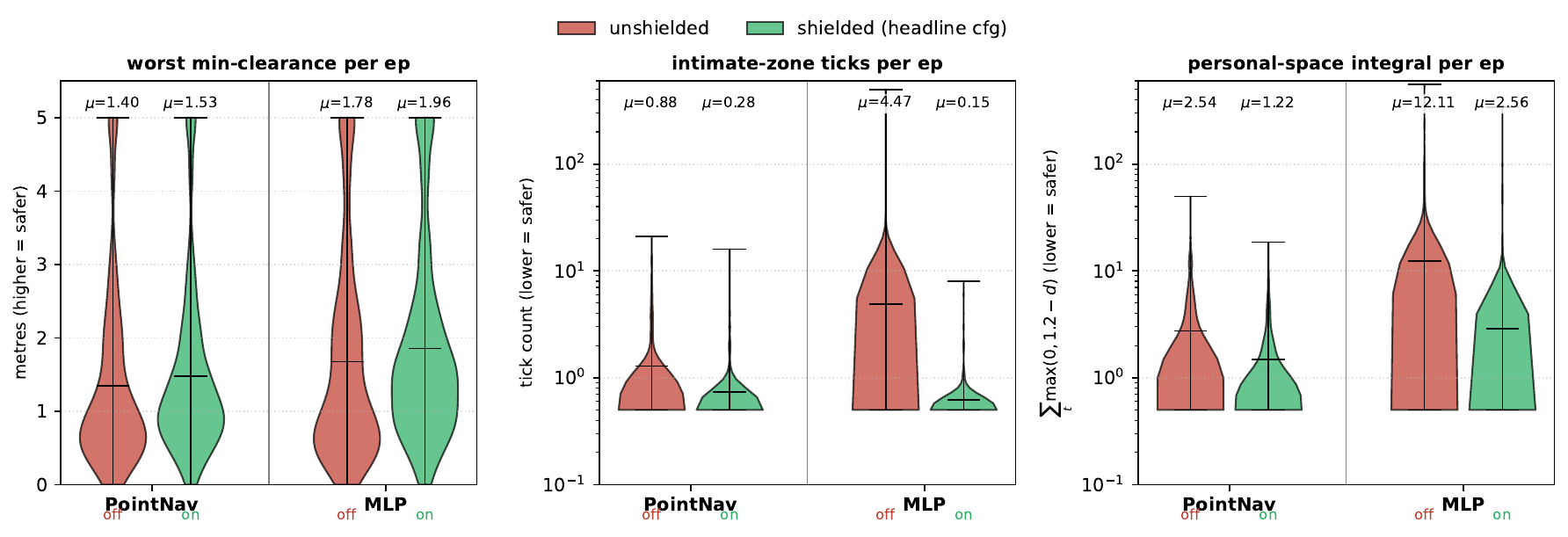}
\caption{Per-episode safety distributions on HA-VLN-CE \texttt{val\_unseen} ($N{=}1839$ per violin) for PointNav and the MLP, unshielded (red) and shielded (green). The shield removes the heavy unsafe tail on both architectures.}
\label{fig:downstream_violins}
\end{figure}

\paragraph{Results.}
On Bench2Drive, the strict shield ($\alpha{=}0$) cuts collisions by $71\%$ and red-light infractions by $77\%$ while raising the Driving Score, and $\alpha{=}0.30$ gives the best Driving Score (Table~\ref{tab:downstream_b2d}; red-light counts and the full $\alpha$ sweep: Table~\ref{tab:downstream_b2d_full}). On HA-VLN-CE, the shield removes $82\%$ of avoidable contacts on PointNav and $93\%$ on the MLP while changing SR by at most $1.1$ points (Table~\ref{tab:downstream_headline}, per-cause breakdown: Appendix~\ref{app:mining_collisions}). A contact is avoidable when the agent was moving forward with the colliding human in its front cone, so a \textsc{stop} would have prevented it. The shield also removes the unsafe tail of the per-episode safety distribution rather than shifting its mean (Fig.~\ref{fig:downstream_violins}). The same recipe yields rules and preferences and transfers zero-shot to new benchmarks and policies.

\subsection{Evolutionary PDDL planning}
\label{sec:exp_planner}

We test whether the planner, using locally-hosted open-weight models, approaches classical planning when the goal vocabulary agrees with the action model and stays robust when it does not.

\paragraph{Setup.} We follow the experimental protocol of NL-PDDL~\citep{liu2026nlpddl}. In open-world ALFWorld\,Text ($N{=}134$ unseen tasks, $50$-step budget), the agent must explore to reveal task objects. Perception and affordances come from the simulator, isolating the planner. In the closed world, the full initial state is given, on the \texttt{alfred} ALFWorld domain and four Blocksworld variants: \emph{Plain}, \emph{Mystery} and \emph{Randomized} (predicates and actions renamed to misleading words or gibberish), and \emph{Misalignment} (goal in plain words, action model renamed). Table~\ref{tab:open_world} reports the open-world results. Closed-world results per variant are in Table~\ref{tab:closed_world} (Appendix~\ref{app:planner_blocksworld}).

\begin{table}[!ht]
\centering
\caption{%
Open-world ALFWorld\,Text ($N{=}134$ unseen tasks), with aligned goals or a goal/action vocabulary mismatch (\textsc{hot}/\textsc{clean}/\textsc{cool} versus \textsc{heated}/\textsc{washed}/\textsc{chilled}). Baseline results are reproduced from~\citep{liu2026nlpddl} (aligned: Table~1, misaligned: Table~3). $^{\dagger}$Token totals are not reported there.
}
\label{tab:open_world}
\scriptsize
\setlength{\tabcolsep}{4pt}
\begin{tabularx}{\linewidth}{l YY YY}
\toprule
 & \multicolumn{2}{c}{Aligned} & \multicolumn{2}{c}{Misaligned} \\
\cmidrule(lr){2-3}\cmidrule(lr){4-5}
Method & \# Tok.\,$\downarrow$ & SR\%\,$\uparrow$ & \# Tok.\,$\downarrow$ & SR\%\,$\uparrow$ \\
\midrule
\multicolumn{5}{l}{\emph{Direct LLM/VLM-based planners}} \\
\quad GPT-4o                                 & 1{,}358{,}934      & 21             & 1{,}440{,}470      & 17 \\
\quad Gemini-2.0 Flash                       & 1{,}176{,}406      & 16             & 1{,}251{,}533      & 15 \\
\quad LLaMA-3.1                              & 1{,}193{,}640      & 19             & 1{,}310{,}664      & 15 \\
\midrule
\multicolumn{5}{l}{\emph{Reflective LLM-based planners}} \\
\quad ReAct (with examples)                  & 5{,}507{,}266      & 81             & 5{,}215{,}612      & 79 \\
\quad ReAct (with model)                     & 4{,}242{,}560      & 34             & 4{,}428{,}914      & 23 \\
\quad Reflexion-3$^{\dagger}$                & NA                 & 83             & NA                 & NA   \\
\quad Reflexion-10$^{\dagger}$               & NA                 & 91             & NA                 & NA   \\
\quad DEPS$^{\dagger}$                       & NA                 & 76             & NA                 & NA   \\
\midrule
BUTLER (fine-tuned, 100K expert traj.)$^{\dagger}$     & NA                 & 26             & NA                 & NA   \\
NL-PDDL~\citep{liu2026nlpddl}                & \textbf{443{,}407} & 94             & \textbf{501{,}049} & 91 \\
\midrule
\ourmethod\ (Qwen3-32B)                      & 853{,}022          & \textbf{98.5}  & 1{,}055{,}341      & \textbf{98.5} \\
\bottomrule
\end{tabularx}
\end{table}

On open-world ALFWorld (Table~\ref{tab:open_world}), \ourmethod\ (Qwen3-32B) solves $98.5\%$ of tasks with both aligned and misaligned goals, above NL-PDDL ($94\%$, $91\%$) and the reflective LLM planners. In the closed world, direct LLM planning collapses on renamed vocabularies ($0\%$ on Mystery and Randomized), whereas \ourmethod\ (Qwen3-32B) reaches $99.7\%$ on Plain, $74.0\%$ on Mystery, $78.7\%$ on Randomized, and $100\%$ on Misalignment, above NL-PDDL ($70\%$) on every variant, and $90.3\%$ on closed-world ALFWorld (Appendices~\ref{app:planner_blocksworld} and~\ref{app:planner_alfworld}).

The planner is not tied to one model. With three locally-hosted open-weight backbones (Table~\ref{tab:backbones}), \ourmethod\ solves $97$--$98.5\%$ of open-world ALFWorld tasks, aligned or misaligned, and $98$--$100\%$ of Blocksworld Plain and Misalignment instances, above NL-PDDL on every benchmark. Gemma~4~31B solves every instance of all four Blocksworld variants.

\begin{table}[!ht]
\centering
\caption{Success rate (\%) of \ourmethod\ with three locally-hosted open-weight backbones on open-world ALFWorld\,Text ($N{=}134$) and the closed-world Blocksworld variants. NL-PDDL numbers are reproduced from~\citep{liu2026nlpddl}.}
\label{tab:backbones}
\small
\setlength{\tabcolsep}{4pt}
\begin{tabularx}{\linewidth}{l YY YYYY}
\toprule
 & \multicolumn{2}{c}{ALFWorld (open world)} & \multicolumn{4}{c}{Blocksworld (closed world)} \\
\cmidrule(lr){2-3}\cmidrule(lr){4-7}
Planner & Aligned & Misaligned & Plain & Mystery & Randomized & Misalign. \\
\midrule
NL-PDDL~\citep{liu2026nlpddl}  & 94   & 91   & 70   & 70   & 70   & 70 \\
\ourmethod\ (Qwen3-32B)         & 98.5 & 98.5 & 99.7 & 74.0 & 78.7 & 100 \\
\ourmethod\ (gpt-oss-120b)      & 97.0 & 97.0 & 100  & 76.7 & 100  & 98 \\
\ourmethod\ (Gemma~4~31B)       & 97.0 & 97.0 & 100  & 100  & 100  & 100 \\
\bottomrule
\end{tabularx}
\end{table}

When the goal vocabulary agrees with the action model, Fast Downward~\citep{helmert2006fastdownward} solves every aligned closed-world task, matching or beating \ourmethod. When it disagrees, Fast Downward is blocked: on misaligned open-world ALFWorld it solves all $59$ tasks whose goal predicates exist in the action model and none of the $75$ heat, clean, and cool tasks whose predicates do not ($44.0\%$) and $0\%$ of Blocksworld Misalignment (Appendix~\ref{app:planner_fd_openworld}). With the predicate map $\mu$ (Sec.~\ref{sec:planning}), \ourmethod\ solves $98$--$100\%$ of these tasks.

Against a single-candidate validate-and-repair loop with the same backbone, budget, and validator, population search solves fewer Mystery instances ($82\%$ and $74\%$ vs.\ $88\%$, Appendix~\ref{app:planner_ablation}). On the $67$ instances both methods solve, population-search plans carry $1.25$ actions above the Fast Downward optimum against $2.24$ ($p{<}0.05$), a $44\%$ reduction in excess plan length.
\subsection{Constrained execution}
\label{sec:exp_endtoend}

We test whether the loop respects $\Phi_{\mathrm{mob}}$ and completes missions in crowded Gazebo scenes and under world changes, then demonstrate it on a Jackal robot.

\begin{figure}[!ht]
\centering
\begin{subfigure}[t]{0.24\linewidth}
  \centering
  \includegraphics[width=\linewidth, height=2.4cm]{\EvoPlanRoot 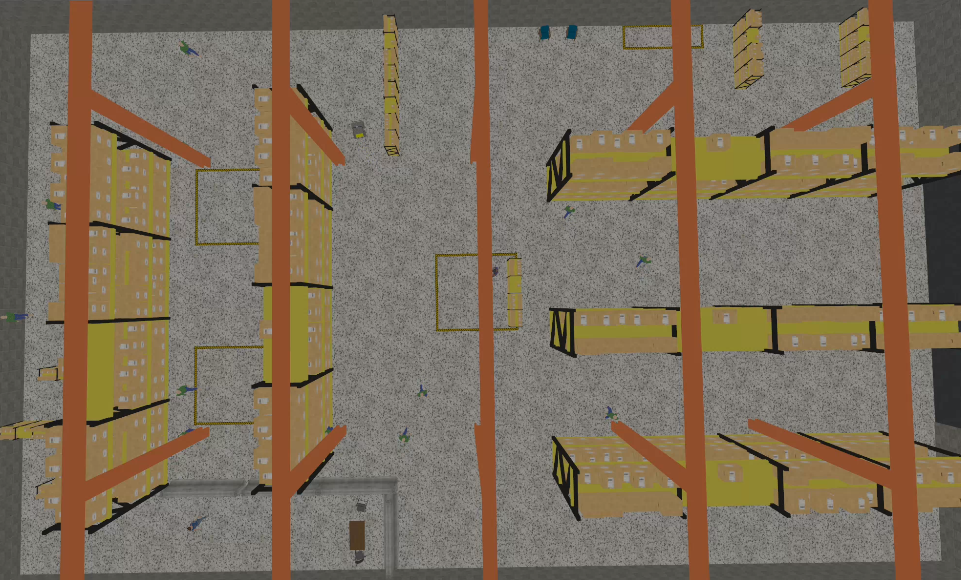}
  \caption{Factory scene}
  \label{fig:gazebo_topdown}
\end{subfigure}\hfill
\begin{subfigure}[t]{0.24\linewidth}
  \centering
  \includegraphics[width=\linewidth, height=2.4cm]{\EvoPlanRoot 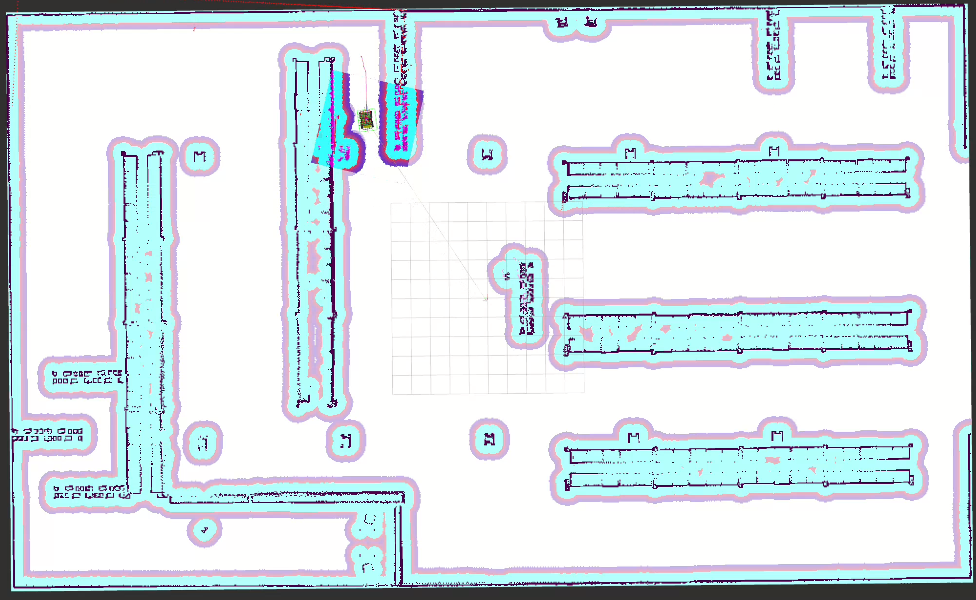}
  \caption{Planner's view}
  \label{fig:rviz_topdown}
\end{subfigure}\hfill
\begin{subfigure}[t]{0.24\linewidth}
  \centering
  \includegraphics[width=\linewidth, height=2.4cm]{\EvoPlanRoot 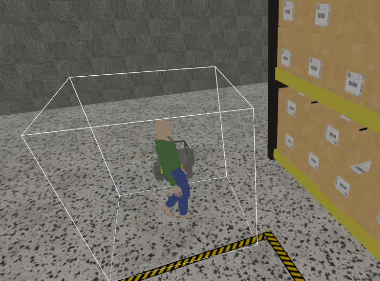}
  \caption{Veto and replan}
  \label{fig:sim_pedestrian}
\end{subfigure}\hfill
\begin{subfigure}[t]{0.24\linewidth}
  \centering
  \includegraphics[width=\linewidth, height=2.4cm]{\EvoPlanRoot 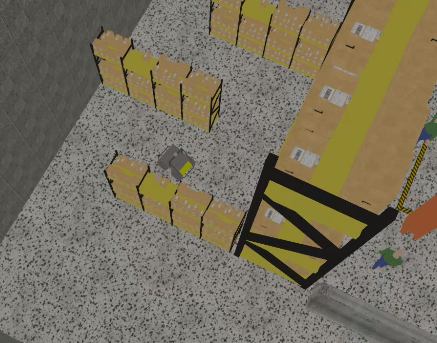}
  \caption{Completed delivery}
  \label{fig:sim_gazebo_aisle}
\end{subfigure}\\[4pt]
\begin{subfigure}[t]{0.32\linewidth}
  \centering
  \includegraphics[width=\linewidth, height=2.6cm]{\EvoPlanRoot 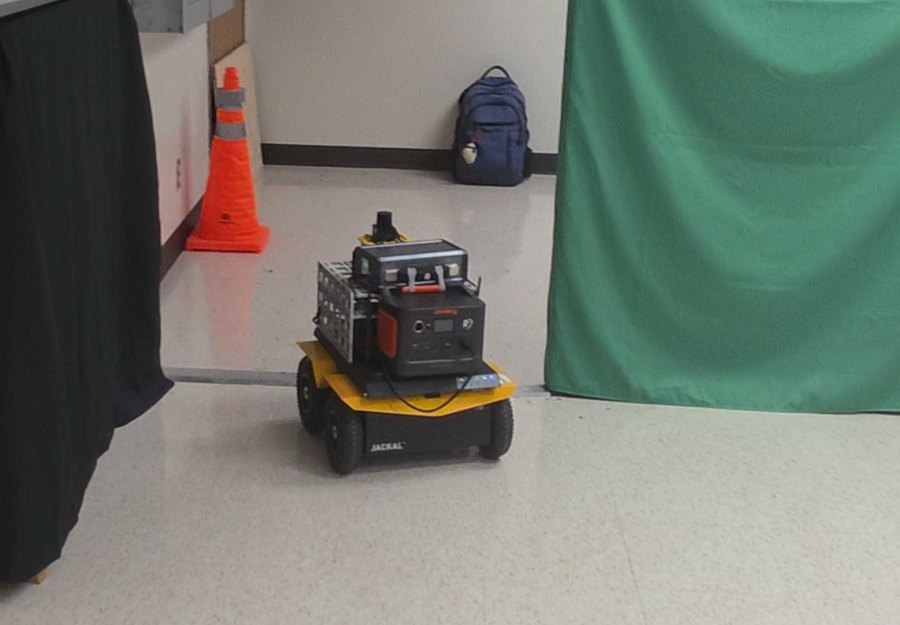}
  \caption{Jackal robot: survey prefix}
\end{subfigure}\hfill
\begin{subfigure}[t]{0.32\linewidth}
  \centering
  \includegraphics[width=\linewidth, height=2.6cm]{\EvoPlanRoot 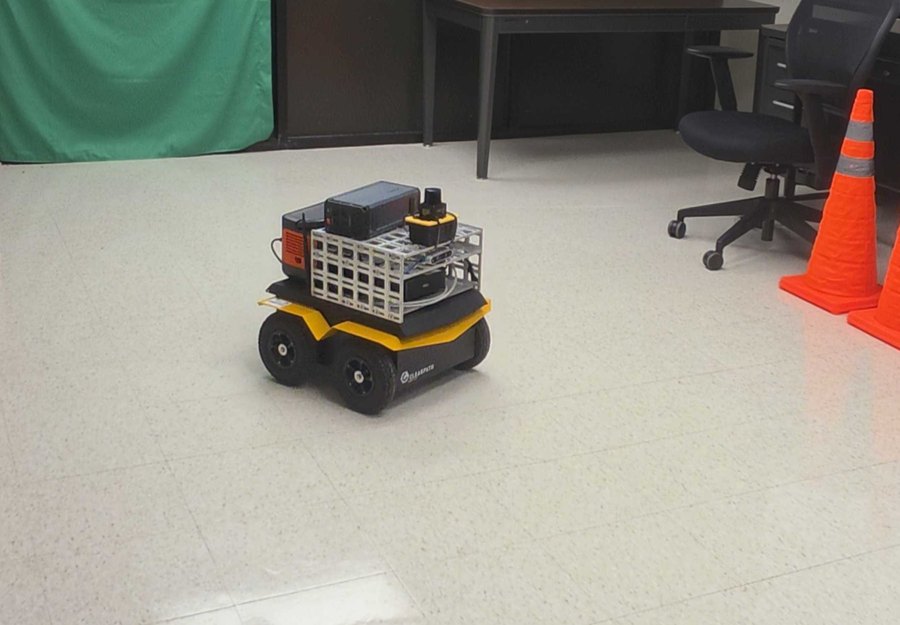}
  \caption{Jackal robot: searching}
\end{subfigure}\hfill
\begin{subfigure}[t]{0.32\linewidth}
  \centering
  \includegraphics[width=\linewidth, height=2.6cm]{\EvoPlanRoot 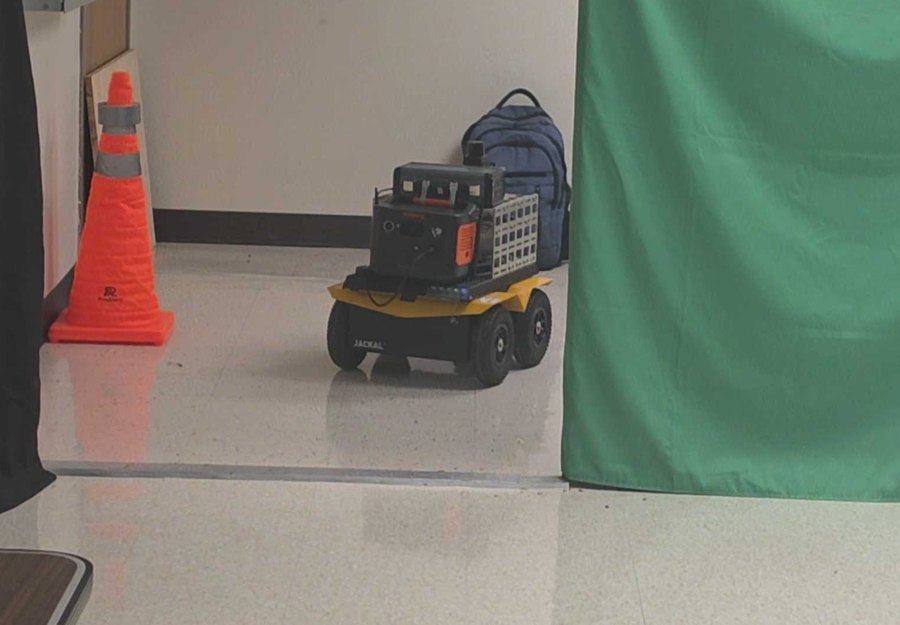}
  \caption{Jackal robot: goal after replan}
\end{subfigure}
\caption{Constrained execution in Gazebo (a--d) and on a Jackal robot (e--g). (b) PDDL \texttt{move} actions become aisle routes. (c) The STL check vetoes a step when pedestrian clearance falls below the learned floor, triggering replanning. (e--g) The backpack's location is unknown, so the robot searches along a VAL-truncated survey prefix, then reaches it after replanning.}
\label{fig:gazebo_pipeline}
\label{fig:real_robot}
\end{figure}

\begin{figure}[!ht]
\centering
\includegraphics[width=\linewidth]{\EvoPlanRoot 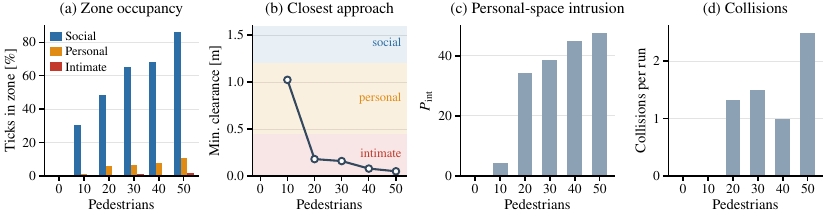}
\caption{Factory crowd sweep ($0$--$50$ pedestrians, $\rho\approx0.0082\,\mathrm{m}^{-2}$ at $10$): (a) zone occupancy, (b) minimum clearance, (c) cumulative personal-space intrusion $P_{\mathrm{int}}$, (d) collisions per run.}
\label{fig:crowd_density_sweep}
\end{figure}

\paragraph{Crowded factory missions.}
A Jackal robot runs $5$ pick-and-drop missions (one run each) among $9$--$15$ walking pedestrians with the SCAND-mined $\Phi_{\mathrm{mob}}$ active (Fig.~\ref{fig:gazebo_pipeline}a--d). Every planned route is completed and four of five missions are collision-free (Table~\ref{tab:factory_missions}). The robot spends $40$--$68\%$ of each mission within social distance ($3.6$\,m) of a pedestrian but at most $13\%$ within personal distance ($1.2$\,m). The one collision (M4, minimum clearance $0.05$\,m) occurs when the densest cluster leaves no admissible replan within the budget. A crowd sweep from $0$ to $50$ pedestrians locates this limit (Fig.~\ref{fig:crowd_density_sweep}): at $10$ pedestrians the loop keeps about $1$\,m minimum clearance with no collisions. Beyond $\rho\approx0.016\,\mathrm{m}^{-2}$ clearance falls into the intimate zone and collisions appear, reaching about $2.5$ per run at $50$ pedestrians, while intimate-zone occupancy stays below $2\%$ of ticks (Appendix~\ref{app:loop_crowd_sweep}).

\begin{table}[!ht]
\centering
\caption{Gazebo missions ($9$--$15$ pedestrians, one run each). Intim/Pers/Social: tick fractions with nearest-pedestrian clearance below $0.45/1.2/3.6$\,m (Hall's zones). $P_{\mathrm{int}}=\sum\max(0,\,1.2-\mathrm{clr})$.}
\label{tab:factory_missions}
\small
\setlength{\tabcolsep}{3pt}
\begin{tabular}{lcccccccccc}
\toprule
Mission & Succ. & Route \% & Coll. & Path [m] & Time [s] & Intim \% & Pers \% & Social \% & $P_{\mathrm{int}}$ & Min clr [m] \\
\midrule
M1 & 1/1 & 100 & 0.0 & 64.1 & 204 & 0.0 &  8.2 & 63.6 &  34.6 & 0.55 \\
M2 & 1/1 & 100 & 0.0 & 90.3 & 241 & 0.5 &  9.8 & 40.1 & 115.2 & 0.40 \\
M3 & 1/1 & 100 & 0.0 & 69.2 & 196 & 0.0 &  4.8 & 67.7 &  12.7 & 0.90 \\
M4 & 0/1 & 100 & 1.0 & 59.1 & 480 & 0.8 &  4.3 & 51.8 &  70.1 & 0.05 \\
M5 & 1/1 & 100 & 0.0 & 37.8 & 112 & 1.0 & 12.7 & 61.8 &  29.4 & 0.37 \\
\bottomrule
\end{tabular}
\end{table}

\paragraph{Replanning missions (Gazebo).}
With YOLO-World perception, a find-an-object mission (one target at an unknown position) completes $4/5$ trials ($227\pm15$\,s, $85.1\pm4.1$\,m), and a find-and-inspect mission that declares no objects in advance, adding each detection to $P$ at run time, completes $3/5$. Failures are Nav2 goal aborts, not invalid plans. In human-avoidance runs with a geometric clearance monitor and a Fast Downward replanner, each run applies two VAL-valid replans after clearance violations and covers all three target regions without collision (Appendix~\ref{app:loop_gazebo_missions}).

\paragraph{Real-world robot experiments.}
On a Jackal robot in a six-region office (Fig.~\ref{fig:gazebo_pipeline}e--g, details: Appendix~\ref{app:loop_real_robot}), the reactive tier re-issues a navigation goal a median $220$\,ms after a clearance violation ($n{=}28$), the deliberative tier returns a VAL-accepted plan in a median $18.0$\,s ($n{=}7$), and the new plan is swapped in under the running executor in a median $3.7$\,ms ($n{=}9$). Across the six complete runs the reactive tier fired $28$ times, and the closest approach, $0.975$\,m inside the $1.00$\,m clearance floor, ended without contact. In these runs the planner is served by gpt-5-mini rather than a locally-hosted model, and a geometric clearance monitor, not $\Phi_{\mathrm{mob}}$, triggers the replans.
\FloatBarrier

\section{Conclusion and Limitations}
\label{sec:conclusion}

The mined $\Phi_{\mathrm{mob}}$, used as a zero-shot shield, cuts collisions by $71\%$ on Bench2Drive and avoidable contacts by $82$--$93\%$ on HA-VLN-CE. The evolutionary planner, on open-weight models, solves $98$--$100\%$ of misaligned tasks where Fast Downward solves $44\%$ and $0\%$. The execution loop completes missions in crowded Gazebo scenes and on a Jackal robot.
\emph{Limitations.} Guarantees are relative to $\Phi_{\mathrm{mob}}$, which covers only the norms our perturbation modes and signal vocabulary represent. When vocabularies agree, Fast Downward is as good or better at lower cost, and population search shortens plans but does not raise success rate over a simple repair loop. Human-avoidance replans used a geometric clearance monitor rather than $\Phi_{\mathrm{mob}}$, and on the real robot each plan repair waits on network latency and LLM inference speed (Appendix~\ref{app:loop_real_robot}).

\clearpage
\acknowledgments{This work was supported in part by NSF grant CNS 2312978. Experiments were conducted on the Texas A\&M University System VISION Cluster and on NVIDIA GPUs provided through the NVIDIA Academic Grant. All opinions expressed are those of the authors and do not necessarily represent those of the sponsoring agencies.}

\bibliography{\EvoPlanRoot references}

\clearpage
\appendix

\begin{tcolorbox}[
  colback=DeepBlue!3, colframe=DeepBlue!60!black, fonttitle=\sffamily\bfseries,
  title={Appendix organisation and supplementary materials},
  boxrule=0.6pt, arc=2pt, left=6pt, right=6pt, top=4pt, bottom=4pt]
The appendix is organised by contribution and mirrors the order of the main paper.
\textbf{Appendix~\ref{app:mining}} expands the offline STL constraint-mining pipeline: trajectory pre-processing, the STL grammar, the evolutionary fitness, and the full downstream-shield results (Bench2Drive rule showcase, HA-VLN-CE preference showcase, per-cause collision decomposition).
\textbf{Appendix~\ref{app:planner}} formalises the EvoPlan algorithm, presents a worked example of plan evolution on a Mystery Blocksworld instance, and reports closed-world Blocksworld and ALFWorld plan-synthesis results together with the same-backbone planner ablation.
\textbf{Appendix~\ref{app:loop}} describes the constrained execution loop -- evolutionary and runtime pseudocode, failure facts, the Nav2 interface, and a crowd-density ablation in the Gazebo factory scene.
\textbf{Appendix~\ref{app:related_extended}} compares related work.

\end{tcolorbox}

\section{Constraint mining: method and validation}
\label{app:mining}

\subsection{Trajectory inputs and preprocessing}
\label{app:mining_inputs}
Each trajectory is converted into a time-indexed record
\begin{equation}
\tau=(x_{0:T},u_{0:T-1},z_{0:T},m_{0:T}),
\end{equation}
where \(x_t\) is robot state, \(u_t\) is control, \(z_t\) is perceptual context, and \(m_t\) contains derived metric signals such as obstacle distance, human distance, terrain label, and region membership. For egocentric videos without reliable pose, structure-from-motion and visual odometry provide approximate trajectories. For SCAND-style data, odometry, joystick commands, lidar, camera, and IMU streams are aligned before mining.

\subsection{Perturbation modes}
\label{app:mining_perturbations}
The perturbed set $\mathcal{D}^-$ is built from the demonstrations themselves. Ten procedural modes (\texttt{speeding}, \texttt{reckless\_turn}, \texttt{crowding\_front}, \texttt{density\_mismatch}, \texttt{red\_light\_run}, \texttt{tailgating}, \dots) each push one or two signals of a demonstration out of its observed range, and an \llm\ generator adds one further perturbation per clip. Each mode encodes only signal \emph{polarity}, i.e.\ which direction of a signal is worse, with no threshold, breakpoint, or conditional structure. The thresholds of $\Phi_{\mathrm{mob}}$ are fitted from the data (App.~\ref{app:mining_negatives}). The grammar, search, and shield are identical for driving rules and social-navigation preferences. Only data and perturbation modes change.

\subsection{STL grammar}
\label{app:mining_grammar}
A compact grammar for the mobility constraint is
\begin{align}
\phi &::= \mu \mid \neg \phi \mid \phi\wedge\phi \mid \phi\vee\phi \mid \Box_{[a,b]}\phi \mid \Diamond_{[a,b]}\phi,\\
\mu &::= d_{obs}\ge c \mid d_{human}\ge c \mid v\le c \mid inside(z) \mid coverage(z)\ge c \mid link\ge c \mid battery\ge c.
\end{align}
In practice the learned $\Phi_{\mathrm{mob}}$ is dominated by global $\Box$ (always) bounds on threshold atoms. The bounded temporal operators $\Box_{[a,b]}$ and $\Diamond_{[a,b]}$ are in the grammar but rarely selected by the evolutionary search on this corpus (App.~\ref{app:mining_prefs_formula} lists all $17$ conjoined $\Box$-bounded clauses).

\subsection{Evolution operators and acceptance tests}
\label{app:mining_evolution}
Formula mutations include threshold perturbation, time-window shift, atom replacement, insertion of $\Box$ or $\Diamond$, and Boolean composition. Crossover swaps subtrees between formulas with compatible signal vocabularies. A formula is accepted when it satisfies four tests: positive held-out robustness on expert traces, low satisfaction rate on perturbed traces, bounded expression complexity, and interpretability by a human reviewer.

\subsection{Mining failure cases}
\label{app:mining_failures}
Common failures include over-specific formulas that memorize a particular route, trivial formulas that are always true, and constraints that depend on unavailable runtime signals. These failures are handled by grammar restrictions, complexity penalties, and signal-availability checks.

\subsection{What the perturbed negatives contribute}
\label{app:mining_negatives}

We test the mining step on the SCAND corpus against two positive-only alternatives and an ablation of our own search (Table~\ref{tab:negatives}). Each constraint is scored on the mining corpus and on real data only: $1{,}696$ simulator-flagged human contacts and $2{,}599$ contact-free windows from unshielded HA-VLN-CE rollouts, episode-disjoint from the mining set.

\begin{table}[!ht]
\centering\small
\caption{Constraints fitted with and without perturbed negatives on the SCAND corpus. TPR is the acceptance rate on demonstrations, Bal.\ acc is balanced accuracy, and Worst is the rejection rate of the weakest perturbation class. Real data: recall on $1{,}696$ human contacts and specificity on $2{,}599$ contact-free windows from HA-VLN-CE.}
\label{tab:negatives}
\begin{tabular}{@{}lccccc@{}}
\toprule
& \multicolumn{3}{c}{Mining corpus} & \multicolumn{2}{c}{Real data} \\
\cmidrule(lr){2-4}\cmidrule(l){5-6}
Fitted from & TPR & Bal.\ acc & Worst & Recall & Spec. \\
\midrule
A.\ Percentile (one-class)            & 0.000 & 0.500          & n/a            & 1.000 & 0.428 \\
B.\ TeLEx~\citep{jha2019telex}         & 0.932 & \textbf{0.828} & 0.026          & 0.499 & 0.788 \\
C.\ Ours (demos $+$ perturbed)        & 0.783 & 0.824          & \textbf{0.457} & 0.518 & 0.777 \\
D.\ Same search, no negatives         & 0.983 & 0.846          & 0.000          & 0.477 & 0.790 \\
\bottomrule
\end{tabular}
\end{table}

\emph{Negatives decide which clauses bind.} Positive-only fits recover threshold values to within $0.02$\,m and match our balanced accuracy, but detection of the weakest perturbation class drops from $0.457$ to $0.026$ (TeLEx) and $0.000$ (same search, no negatives).
\emph{Negatives do not set the constants.} Re-mining across a $16\times$ sweep of perturbation magnitude moves the learned thresholds by a median of $0.6\%$, and the clearance floor stays at \texttt{min\_clearance} $\geq 0.480$\,m. Where the generator does fix a constant (\texttt{crowd\_count} $\geq 6$), the miner selects a lower $3$.
\emph{Coverage is limited to the modelled taxonomy.} Withholding all modes of one class retains $53\%$ (clearance) and $57\%$ (social) of the perturbation-attributable detection, where demonstrations alone already suffice, but only $4\%$ (density) and $10\%$ (kinematic), where the perturbations carry the detection.
\emph{Vocabulary sensitivity is class-specific.} The signal vocabulary is pre-specified but proposed by an LLM. Dropping the pedestrian channels lowers social detection from $0.90$ to $0.33$ while clearance stays at $0.97$. Dropping \texttt{crowd\_count} lowers density detection from $0.76$ to $0.36$. Conditional guards add $0.45$ density detection with the full vocabulary but only $0.04$ without a context signal, and an $819$-candidate algebraic search over derived signals gains only $0.001$.

\subsection{Rule showcase: Bench2Drive shield setup and full results}
\label{app:mining_rules}

\paragraph{Recovered formula.}
The nuPlan extraction recovers a small set of conditional implications, each with binary right/wrong semantics: a stop-on-red-light implication, vehicle-clearance and time-to-collision floors, and lane-aware speed caps. The perturbed negatives that falsify these clauses (a \texttt{red\_light\_run} or a closing-rate spike inside the lead-vehicle gap) are unambiguous rule violations rather than envelope-tightness perturbations.

\paragraph{Keyword interface and continuous expansion.}
The Qwen2.5-VL agent is prompted every ten simulator ticks ($\approx\!1$\,Hz) on the forward-camera frame and emits one of four keywords $\{\textsc{accelerate}, \textsc{coast}, \textsc{brake}, \textsc{emergency\_stop}\}$. A latched target-speed updated by each keyword (e.g., \textsc{accelerate} raises it by $\Delta v$, \textsc{brake} halves it) and a per-tick P-controller produce continuous (throttle, steer, brake) commands to CARLA on every intervening tick. The policy's effective control authority is therefore continuous even though the shield's decision space is the four-option discrete keyword set.

\paragraph{Shield loop.}
The shield wraps the keyword decision identically to the HA-VLN-CE case of App.~\ref{app:mining_prefs_shield}: predict the short-horizon rollout for each candidate keyword, compute min-robustness $\rho$ against the mined formula, veto when $\rho{<}0$ under an imminence gate, and substitute via MPC over the four keywords with the policy-prior weight $\alpha$.

\paragraph{Full Bench2Drive sweep.}
Table~\ref{tab:downstream_b2d_full} adds the full policy-prior sweep on Bench2Drive (all six shielded $\alpha$ settings) with three additional metrics (route completion, total vetoes, completed routes) on top of the condensed main-text Table~\ref{tab:downstream_b2d}.

\begin{table}[!ht]
\centering
\caption{Full nuPlan-mined rule shield results on Bench2Drive across the policy-prior sweep. The columns report Driving Score (DS), Route Completion (RC), collisions (Coll), red-light infractions (RL), total infractions (Infr), total shield substitutions (Vetoes), and completed routes (Succ). Best value in each column is in bold.}
\label{tab:downstream_b2d_full}
\small
\begin{tabularx}{\linewidth}{l Y Y Y Y Y Y Y}
\toprule
Config & DS\,$\uparrow$ & RC\,$\uparrow$ & Coll\,$\downarrow$ & RL\,$\downarrow$ & Infr\,$\downarrow$ & Vetoes & Succ \\
\midrule
Unshielded                       & 62.96 & \textbf{69.2} & 14         & 13         & 20         & 0        & \textbf{32} \\
Shielded, $\alpha{=}0$ (strict)  & 65.61 & 67.2          & \textbf{4} & \textbf{3} & 7          & 7{,}784  & 30 \\
Shielded, $\alpha{=}0.05$        & 65.10 & 67.3          & 5          & 6          & 9          & 7{,}807  & 30 \\
Shielded, $\alpha{=}0.10$        & 65.37 & 67.6          & 5          & 6          & 9          & 8{,}047  & 31 \\
Shielded, $\alpha{=}0.20$        & 65.07 & 67.9          & 9          & 5          & 12         & 7{,}041  & 31 \\
Shielded, $\alpha{=}0.30$        & \textbf{66.20} & 67.8 & \textbf{4} & 4          & \textbf{6} & 8{,}552  & 31 \\
Shielded, $\alpha{=}0.50$        & 64.85 & 67.6          & 9          & 7          & 13         & 10{,}157 & 31 \\
\bottomrule
\end{tabularx}
\end{table}

\subsection{Preference showcase: architecture, formula, shield mechanism, $\alpha$ analysis}
\label{app:mining_prefs}

\paragraph{Base policies.}
PointNav is the pretrained PointNavResNet50$+$LSTM policy from DD-PPO~\citep{wijmans2019ddppo}, evaluated on RGB+depth observations from the HA-VLN-CE simulator. The MLP is a stateless two-hidden-layer network ($256{\times}256$, ReLU) trained by behaviour cloning on the PointNav policy's own decisions across the \texttt{train} split, so it represents a memoryless approximation of the same input--output mapping. Both policies emit one of four discrete actions $\{\textsc{stop}, \textsc{fwd}, \textsc{left}, \textsc{right}\}$ per tick directly to the simulator.

\paragraph{Episode protocol and metrics.}
Each condition is evaluated on the full HA-VLN-CE \texttt{val\_unseen} split, $N{=}1839$ episodes, paired by episode ID so the unshielded and shielded conditions experience the same humans, layouts, and goal poses. An episode is a \emph{collision-termination} (CT) when the simulator's collision counter increments while \texttt{min\_clearance}${<}1.0$\,m. A \emph{success} is reached-goal-and-never-human-collision. We additionally classify each collision-terminated episode by its final-tick safety log into \emph{avoidable contacts} (the agent walked into a stationary or slowly-moving human) and \emph{trajectory-crossings} (a side-cone human only became visible inside the $K$-step shield rollout). The per-cause decomposition appears in App.~\ref{app:mining_collisions}.

\paragraph{Full mined STL formula.}
\label{app:mining_prefs_formula}
The OpenEvolve~\citep{openevolve} mining loop converges to the conjunction of $17$ globally-quantified predicates below. Thresholds are fitted by evolutionary search, not hand-set.

\begin{lstlisting}[style=compactcode]
G(speed              <= 2.603) &
G(yaw_rate           <= 3.293) &
G(accel              <= 3.213) &
G(jerk               <= 5.862) &
G(lat_accel          <= 2.228) &
G(cmd_steer          <= 1.000) &
G(min_clearance      >= 0.483) &
G(crowd_count >= 3 | min_clearance >= 0.6) &
G(front_clearance    >= 0.505) &
G(left_clearance     >= 0.500) &
G(right_clearance    >= 0.500) &
G(rear_clearance     >= 0.483) &
G(ttc                >= 0.452) &
G(ped_approach_rate  <= 5.000) &
G(ped_clearance      >= 0.500) &
G(front_ped_clearance>= 0.500) &
G(ped_ttc            >= 0.422)
\end{lstlisting}

Three clause families are present. The \emph{kinematic comfort envelope} (six clauses) bounds \texttt{speed}, \texttt{yaw\_rate}, \texttt{accel}, \texttt{jerk}, \texttt{lat\_accel}, and operator \texttt{cmd\_steer}. The \emph{social keep-distance floors} (seven clauses) bound every directional clearance bin and the time-to-collision, and add the crowd-conditioned disjunction \texttt{G(crowd\_count}${\ge}$3\texttt{ | min\_clearance}${\ge}$0.6\texttt{)} that catches the \texttt{density\_mismatch} perturbed negatives the flat \texttt{min\_clearance} floor misses. The \emph{pedestrian-specific clauses} (four clauses) mirror the same structure on \texttt{ped\_clearance} signals available when perception has separated pedestrians from generic returns. The $0.5$\,m floor on \texttt{front\_ped\_clearance} is the binding clause at downstream shield time.

\paragraph{Shield mechanism.}
\label{app:mining_prefs_shield}
At each control tick the shield performs the following steps.
\begin{enumerate}
\item Extract a current signal dictionary from the live simulator (the same clearance bins SCAND uses, computed without perception because HA-VLN exposes ground-truth human positions).
\item Call the base agent's \texttt{act(obs)} to get a proposed action $a_t$.
\item Hypothesise the next-tick signals under $a_t$, append them to a $10$-tick rolling window, and evaluate the formula's min-robustness $\rho$.
\item If $\rho<0$ \emph{and} the minimum predicted clearance over the $K$-step rollout is below an \emph{imminence threshold}, the action is \textbf{vetoed}. This threshold is $1.0$\,m for PointNav and the MLP.
\item On veto, a $K{=}5$-step model-predictive substitution enumerates the four discrete actions, rolls each forward under constant-velocity human motion and analytic agent kinematics, scores by predicted $\rho$, and commits the argmax first action. An optional beam variant (width $B{=}4$) plans over action sequences of length \texttt{mpc\_horizon}. It finds recovery patterns such as ``stop, turn, forward'' that the single-step substitution misses.
\end{enumerate}

\paragraph{Predictive imminence.}
An early version of the shield gated only on the next-tick \texttt{front\_ped\_clearance}. Side humans about to cross the agent's path $2$--$5$ ticks ahead kept this signal high at tick $1$, so the shield silently let the agent walk into the crossing. This trajectory-crossing failure mode dominated $65\%$ of the residual collisions. The current shield gates on the minimum clearance over the full $K$-step rollout, taken across \texttt{min\_clearance} (any direction), \texttt{front\_ped\_clearance}, and \texttt{front\_obstacle\_clearance}. The any-direction component is the critical addition: side humans that do not show up in the front cone do show up in \texttt{min\_clearance}.

\paragraph{Policy-prior weight $\alpha$.}
A soft policy prior augments the candidate score by the log-probability the base policy assigned to each action:
\begin{equation}
\hat a = \operatorname*{arg\,max}_{a}\; \rho(\tau_a) + \alpha\,\log\bigl(\max(\epsilon, \Pr_\pi[a])\bigr),
\end{equation}
with $\epsilon{=}10^{-2}$ flooring the log to keep the bonus bounded. At $\alpha{=}0$ the prior is disabled and the picker recovers the original min-robustness argmax. With the floor, the maximum penalty for a zero-probability action is $\alpha\log(10^{-2})\approx -4.6\alpha$, so for $\alpha\in\{0.05, 0.1, 0.2\}$ the prior caps at $\{0.23, 0.46, 0.92\}$\,m of $\rho$-equivalent: a meaningful tiebreak among comparable-safety candidates that never overrides a ${>}1$\,m clearance differential. On PointNav, $\alpha{=}0.05$ is the only sweep value that does not over-trust the policy ($\alpha\in\{0.1,0.2\}$ pushes the veto rate to $6$--$7\%$ and degrades both SR and CT). On the stateless MLP, $\alpha{=}0$ already preserves SR within noise and no prior is needed.

\paragraph{Why the prior is necessary: preserving PointNav's training distribution.}
PointNav is a DD-PPO-trained recurrent policy~\citep{wijmans2019ddppo}: its LSTM hidden state encodes recent actions through a learned \texttt{prev\_action} embedding, and the policy is almost deterministic given an observation. Without the prior, the safety picker substitutes whenever any candidate has higher predicted $\rho$, even by a margin much smaller than the typical safety differential. Each such substitution drives the LSTM off-distribution at every subsequent tick: the hidden state has ``just moved forward'' encoded in it while the simulator actually advanced under \textsc{stop} or a turn, so the visual encoder sees paths DD-PPO never traversed during training. The prior fixes this by requiring a substitution's safety advantage to exceed the policy's own confidence in its proposal. The bounded $-4.6\alpha$ floor keeps this soft: a substantial $\rho$ gap (${>}1$\,m clearance for $\alpha{=}0.05$) still overrides the prior, so the prior only suppresses small-margin substitutions that would have desynced the policy without buying real safety. The stateless MLP needs no prior because it has no recurrent state to corrupt --- which is why $\alpha{=}0$ preserves SR on MLP (Table~\ref{tab:downstream_headline}) but costs $4.3$\,pp on PointNav, and $\alpha{=}0.05$ is the smallest nonzero setting that closes that gap.

\paragraph{Full HA-VLN-CE results with confidence intervals.}
Table~\ref{tab:downstream_headline_full} adds the McNemar paired $95\%$ confidence intervals on $\Delta$SR and $\Delta$CT for HA-VLN-CE.

\begin{table}[!ht]
\centering
\caption{Full SCAND-mined preference shield results on HA-VLN-CE \texttt{val\_unseen}, $N{=}1839$ paired episodes per row. CIs use McNemar paired SE on discordant cells, and ``ns'' marks $|z|{<}2$.}
\label{tab:downstream_headline_full}
\small
\begin{tabularx}{\linewidth}{l l Y Y Y Y}
\toprule
Policy & Shield & SR\%\,$\uparrow$ & CT\%\,$\downarrow$ & $\Delta$SR (95\% CI) & $\Delta$CT (95\% CI) \\
\midrule
PointNav         & off & 58.7 & 38.1 & - & - \\
\quad\emph{$\alpha{=}0.05$} & on  & 57.6 & \textbf{18.5} & $-1.1$\,pp $[-2.9,+0.7]$ ns & $\mathbf{-19.6}$\,pp $[-22.0,-17.2]$ \\
MLP              & off & 32.1 & 30.1 & - & - \\
\quad\emph{$\alpha{=}0$}    & on  & 32.7 & \textbf{5.2}  & $+0.6$\,pp $[-0.7,+2.0]$ ns & $\mathbf{-24.9}$\,pp $[-27.2,-22.6]$ \\
\bottomrule
\end{tabularx}
\end{table}

\subsection{Per-cause collision decomposition}
\label{app:mining_collisions}

We classify each collision-terminated episode using the $5$-tick safety log captured immediately before termination. An \emph{avoidable} contact is one where, in the final tick, the agent was moving forward (\texttt{speed}${>}0.1$) and the colliding human was in the front cone (\texttt{front\_clearance}${<}1.5$\,m), so a \textsc{stop} would have prevented the contact. A \emph{trajectory-crossing} contact is one where the agent was moving but the colliding human was \emph{not} in the front cone. The human's path crossed the agent's from the side or rear, leaving only the predictive-rollout horizon to react. A third bucket, \emph{human-into-static-agent}, is always empty in our data because HA-VLN's \texttt{previous\_step\_collided} flag only fires when the agent attempts a move that is physically blocked.

\begin{table}[!ht]
\centering
\caption{Per-cause breakdown of collision-terminated episodes (out of $1839$ each), on the headline configurations of Table~\ref{tab:downstream_headline}.}
\label{tab:app_downstream_avoidable}
\small
\begin{tabularx}{\linewidth}{l l Y Y Y Y}
\toprule
Policy & Cause & unshielded\,$\downarrow$ & shielded\,$\downarrow$ & $\Delta$\,$\downarrow$ & cut \\
\midrule
\textbf{PointNav} & agent-walked-into (avoidable) & 241 (13.1\%) & \textbf{44 (2.4\%)}  & $-197$ & $\mathbf{-82\%}$ \\
                  & trajectory-crossing (mixed)    & 455 (24.7\%) & 297 (16.2\%)         & $-158$ & $-35\%$ \\
                  & human-into-static (unavoid.)   & 0            & 0                    & -    & -     \\
                  & \emph{total CT}                & 696 (37.8\%) & 341 (18.5\%)         & $-355$ & $-51\%$ \\
\midrule
\textbf{MLP}      & agent-walked-into (avoidable) & 209 (11.4\%) & \textbf{14 (0.8\%)}  & $-195$ & $\mathbf{-93\%}$ \\
                  & trajectory-crossing (mixed)    & 344 (18.7\%) & 81 (4.4\%)           & $-263$ & $\mathbf{-76\%}$ \\
                  & human-into-static (unavoid.)   & 0            & 0                    & -    & -     \\
                  & \emph{total CT}                & 553 (30.1\%) & 95 (5.2\%)           & $-458$ & $\mathbf{-83\%}$ \\
\bottomrule
\end{tabularx}
\end{table}

Two takeaways from Table~\ref{tab:app_downstream_avoidable}.\footnote{Per-cause counts are from a separate evaluation of the same configuration. Over five repeated evaluations, unshielded PointNav ends in a collision termination in $38.34\pm0.19\%$ of episodes.} First, the shield is near-perfect on its core jurisdiction: avoidable contacts drop by ${\ge}82\%$ on both policies, the cleanest demonstration that the mined constraint does what its name promises. Second, the shield also helps on trajectory-crossings ($-35\%$ on PointNav, $-76\%$ on MLP) because the $K$-step predictive rollout extrapolates side humans' positions forward and vetoes a \textsc{forward} that would put the agent in a future contact, even when the human is not yet in the front cone at decision time.

\section{Evolutionary PDDL planner: algorithm and results}
\label{app:planner}

\subsection{EvoPlan algorithm}
\label{app:planner_algorithm}

\paragraph{Fitness.} The cascade scores a well-formed candidate plan $\pi$ as
\begin{equation}
\label{eq:fitness}
J(\pi) =
\begin{cases}
1 & \text{if VAL accepts } \pi,\\
0.1 + 0.7\,k/|\pi| & \text{otherwise,}
\end{cases}
\end{equation}
where $k$ is the number of actions executed before the first failure, so valid plans always dominate and invalid ones are ranked by their executable prefix.

\paragraph{Search details.} Typical LLM edits insert a missing precondition action, repair an object binding, swap independent actions, or resample a suffix, but the edit type is not constrained in advance. When continuous execution is involved, the cascade adds a fourth stage that scores the predicted robustness of the plan's motions against $\Phi_{\mathrm{mob}}$. Invalid plans are demoted by the cascade, not by an LLM critic. Validator feedback names only the failing action, so each prompt targets the plan's unverified part.

EvoPlan combines an \emph{outer} task loop driven by belief growth in the simulator with an \emph{inner} evolutionary loop that proposes and repairs PDDL plans. The outer loop alternates between exploring to expand the belief and calling the inner loop on the resulting PDDL problem. The inner loop mutates an LLM-proposed plan against feedback from a symbolic validator cascade. When the goal vocabulary disagrees with the action-model vocabulary (misalignment), a one-time LLM call derives a predicate-symbol map $\mu$ that is cached and reused. The algorithm consumes a fixed PDDL domain $D$ provided as input.

The notation used in Algorithm~\ref{alg:evoplan} is summarised first. Symbols are the variables that flow through the procedures. Routines are the named subprocedures (their bodies are described informally rather than expanded). For clarity, Algorithm~\ref{alg:evoplan} serves as the formal description of EvoPlan, and the accompanying Tables of Symbols and Routines provide a convenient reference for notation and procedures.
\paragraph{Symbols (state and inputs).}
\begin{description}[leftmargin=2.4em, itemsep=1pt, topsep=2pt, font=\normalfont\itshape]
\item[$D, g$] PDDL domain and goal predicates (inputs).
\item[$M$] LLM backbone (used for proposal, repair, and the predicate map $\mu$).
\item[$E$, $B_{\max}$] TextWorld simulator and total simulator-step budget.
\item[$\mathcal{B}$] running belief: known receptacles plus observed objects.
\item[$P$, $\pi$] PDDL problem instance and grounded action sequence.
\item[$\mu$] plain$\to$misaligned predicate-symbol map (\textsc{nil} if vocabularies agree).
\item[$\mathrm{pop}$, $c$, $f_i$, $J$] population, candidate, cascade feedback at iteration $i$, scalar fitness of Eq.~\eqref{eq:fitness}.
\end{description}

\paragraph{Routines (subprocedure semantics).}
\begin{description}[leftmargin=2.4em, itemsep=1pt, topsep=2pt, font=\normalfont]
\item[\textsc{target\_ready}$(\mathcal{B},g)$] true iff every goal-relevant object and affordance is already in $\mathcal{B}$.
\item[\textsc{belief\_to\_problem}$(\mathcal{B},D,g)$] emits PDDL with the goal existential over the target type.
\item[\textsc{pick\_receptacle}$(\mathcal{B},g,M)$] selects an unvisited receptacle. In \emph{targeted} mode the LLM ranks receptacle types by likelihood of holding $g$'s target.
\item[\textsc{cascade}$(c,D,P)$] runs syntax $\to$ VAL ($\to$ STL robustness when a continuous trace exists) and returns $(\text{valid},\text{feedback})$.
\item[\textsc{rewrite\_heads}$(P,\mu)$] regex substitution of predicate heads in $P$ by $\mu$.
\end{description}

\begin{algorithm}[!ht]
\caption{EvoPlan. The outer procedure \textsc{Plan} drives belief-based exploration and the inner \textsc{Evolve} performs LLM-guided plan repair against the cascade. The predicate map $\mu$ is derived once per task on a vocabulary mismatch and cached to disk.}
\label{alg:evoplan}
\begin{algorithmic}[1]
\Procedure{Plan}{$D, g, M, B_{\max}, E$}
    \State $\mathcal{B} \gets E.\textsc{reset}()$;\quad $t \gets 0$;\quad $\text{won} \gets \text{false}$
    \While{$t < B_{\max}$ \textbf{and not} won}
        \If{$\textsc{target\_ready}(\mathcal{B}, g)$}
            \State $P \gets \textsc{belief\_to\_problem}(\mathcal{B}, D, g)$
            \State $\pi \gets \Call{Evolve}{D, P, M, g}$
        \Else
            \State $\pi \gets \emptyset$ \Comment{nothing to plan for, so trigger exploration}
        \EndIf
        \If{$\pi = \emptyset$} \Comment{exploration step}
            \State $r \gets \textsc{pick\_receptacle}(\mathcal{B}, g, M)$
            \State $\pi \gets \langle \text{goto}(r) \rangle$ extended by $\langle \text{open}(r) \rangle$ if $r$ is openable
        \EndIf
        \For{$a \in \pi$ \textbf{while} $t < B_{\max}$ \textbf{and not} won}
            \State $(o, \text{won}) \gets E.\textsc{step}(a)$;\quad $t \gets t + 1$
            \State $\mathcal{B} \gets \textsc{update\_belief}(\mathcal{B}, o, a)$
        \EndFor
    \EndWhile
    \State \Return $\text{won}$
\EndProcedure
\Statex
\Procedure{Evolve}{$D, P, M, g$}
    \If{$\mathrm{vocab}(g) \not\subseteq \mathrm{vocab}(D)$}
        \State $\mu \gets$ \Call{map\_predicates}{$g, D, M$} \Comment{one LLM call per task, disk-cached}
        \State $P \gets \textsc{rewrite\_heads}(P, \mu)$
    \EndIf
    \State $\mathrm{pop} \gets \{\textsc{seed}(D, P)\}$
    \For{$i = 1, \dots, I$}
        \State $c \gets M.\textsc{mutate}\bigl(\textsc{select}(\mathrm{pop}),\, f_{i-1}\bigr)$
        \State $(\text{valid}_i, f_i) \gets \textsc{cascade}(c, D, P)$
        \State $\mathrm{pop}.\textsc{add}(c, J(c, f_i))$
        \If{$\text{valid}_i$} \Return $c.\pi$ \EndIf \Comment{early exit on first valid plan}
    \EndFor
    \State \Return $\emptyset$ \Comment{no valid plan within the inner budget}
\EndProcedure
\end{algorithmic}
\end{algorithm}

\paragraph{Worked example.} Algorithm~\ref{alg:evoplan} run on Blocksworld \emph{Mystery} instance \texttt{p0009} (action names renamed to deceptive English, with the goal of stacking yellow-on-pink-on-orange) converges in four iterations. Each row pairs a candidate plan with validator feedback for the next mutation.

\begin{tcolorbox}[
  breakable,
  colback=DeepBlue!3, colframe=DeepBlue!60!black, fonttitle=\sffamily\bfseries,
  title={Evolution of an EvoPlan candidate on Mystery Blocksworld instance \texttt{p0009}},
  boxrule=0.6pt, arc=2pt, left=6pt, right=6pt, top=4pt, bottom=4pt]
\footnotesize

\textbf{\textsc{Iter 1}} \hfill
\colorbox{red!55!black}{\color{white}\sffamily\bfseries\,INVALID\,} \enspace
\textsf{steps $3/10$, score $0.31$}
\\[2pt]
{\small\ttfamily
(feast orange pink) (succumb orange) (attack pink) \textcolor{red!70!black}{\textbf{(attack orange)$\,\leftarrow$ FAIL}} (succumb orange) (overcome pink orange) (feast yellow green) (succumb yellow) (attack yellow) (overcome yellow pink)
}
\\[1pt]
\textsf{Feedback $\to$ next iter:} \emph{(attack orange) has an unsatisfied precondition at time 4 -- set (harmony) to true.}

\smallskip
\noindent\hfil$\downarrow$\hfil

\smallskip
\textbf{\textsc{Iter 2}} \hfill
\colorbox{orange!65!black}{\color{white}\sffamily\bfseries\,INVALID\,} \enspace
\textsf{steps $16/16$, score $0.80$}
\\[2pt]
{\small\ttfamily
(feast orange pink) (succumb orange) (attack pink) (succumb pink) (feast yellow green) (succumb yellow) (attack yellow) (succumb yellow) (attack yellow) (overcome yellow pink) (feast yellow pink) (succumb yellow) (attack pink) (succumb pink) (attack pink) (overcome pink orange)
}
\\[1pt]
\textsf{Feedback $\to$ next iter:} \emph{the goal is not satisfied -- set (craves yellow pink) to true.}

\smallskip
\noindent\hfil$\downarrow$\hfil

\smallskip
\textbf{\textsc{Iter 3}} \hfill
\colorbox{yellow!75!black}{\color{white}\sffamily\bfseries\,INVALID\,} \enspace
\textsf{steps $9/14$, score $0.55$}
\\[2pt]
{\small\ttfamily
(feast orange pink) (succumb orange) (attack pink) (overcome pink orange) (feast yellow green) (succumb yellow) (attack yellow) (overcome yellow pink) (feast yellow pink) \textcolor{red!70!black}{\textbf{(attack pink)$\,\leftarrow$ FAIL}} (succumb pink) (attack pink) (succumb pink) (overcome pink orange)
}
\\[1pt]
\textsf{Feedback $\to$ next iter:} \emph{(attack pink) has an unsatisfied precondition at time 10 -- set (harmony) to true.}

\smallskip
\noindent\hfil$\downarrow$\hfil

\smallskip
\textbf{\textsc{Iter 4}} \hfill
\colorbox{DeepGreen!65!black}{\color{white}\sffamily\bfseries\,\checkmark\ VALID\,} \enspace
\textsf{steps $8/8$, score $1.00$}
\\[2pt]
{\small\ttfamily
(feast orange pink) (succumb orange) (attack pink) (overcome pink orange) (feast yellow green) (succumb yellow) (attack yellow) (overcome yellow pink)
}
\\[1pt]
\textsf{Loop terminates:} \emph{the plan is valid (final value $8$). No further iteration is needed.}
\end{tcolorbox}

The pattern -- precondition fail, goal-miss over-correction, partial repair with spurious tail, then a clean trim to the optimum -- is typical: the validator artifact converts ``which clause failed'' into the next prompt's repair target.

\subsection{Closed-world Blocksworld results}
\label{app:planner_blocksworld}

This section evaluates EvoPlan on the closed-world benchmarks of the NL-PDDL suite~\citep{liu2026nlpddl}, where the planning component is isolated from perception and exploration. The only LLM call in the loop is plan proposal/repair. Validity is decided by \val and Fast Downward~\citep{howey2004val,helmert2006fastdownward}.

\paragraph{Benchmark variants.} Following~\citet{liu2026nlpddl}, we use the standard Blocksworld domain~\citep{mcdermott1998pddl} (\emph{Plain}) and three surface-form variants that defeat lexical priors: \emph{Mystery} (predicates and actions renamed to misleading English words, introduced in PlanBench~\citep{valmeekam2023planbench}), \emph{Randomized} (renamed to gibberish), and \emph{Misalignment} (action model in renamed terms, goal in plain vocabulary). On Misalignment, one LLM call derives the predicate map $\mu$ (Sec.~\ref{sec:planning}) that rewrites the goal predicates into the misaligned vocabulary. The evolutionary loop runs unchanged thereafter.

\paragraph{Headline result.} Table~\ref{tab:closed_world} reports SR per variant. Direct LLM planners collapse on the renamed variants ($0\%$ on Mystery/Randomized). EvoPlan recovers near-symbolic performance: both backbones reach $99.7$--$100\%$ on \emph{Plain}, \emph{gpt-oss-120b} matches FD on \emph{Randomized}, and both backbones exceed NL-PDDL on every variant. On \emph{Misalignment} (FD $= 0\%$), we reach $98$--$100\%$ using $\mu$.

\paragraph{Mystery as the difficulty floor.} \emph{Mystery} is the hardest variant for Qwen3-32B and exposes a backbone-specific limit: the deceptive vocabulary inflates the model's reasoning trace, and clearing NL-PDDL's $70\%$ requires raising the per-call reasoning budget. On \emph{gpt-oss-120b} the same variant is reached at a lower budget, indicating that the bottleneck is per-call reasoning depth rather than the number of iterations. Across all variants, most tasks converge within $1$--$3$ iterations and per-call prompts carry no few-shot exemplars.

\paragraph{Reporting conventions.} The benchmark is the NL-PDDL Blocksworld suite~\citep{liu2026nlpddl}: Plain plus the three surface-form variants Mystery, Randomized, and Misalignment. Tokens-to-solve are computed as the cumulative LLM tokens up to the iteration that first emits a VAL-valid plan. Unsolved tasks count their full iteration budget. Direct-LLM, NL-PDDL, and Fast Downward~\citep{helmert2006fastdownward} numbers are reproduced from~\citet{liu2026nlpddl}. NL-PDDL's token cells are dashed on Plain/Mystery/Randomized because its FOL regression planner makes no LLM calls when the goal and action-model vocabularies agree (\citet{liu2026nlpddl} report $0$ tokens). On Misalignment it issues repeated commonsense-entailment queries during regression, reusing a cache of earlier answers, and its token cell reflects these calls. On Misalignment, \ourmethod\ analogously derives $\mu$ with one LLM call before evolution.

\begin{table}[!ht]
\centering
\caption{Closed-world Blocksworld plan-synthesis: tokens-to-solve (\# Tok., per $N{=}100$ instances) and success rate (SR) across the four NL-PDDL variants. Best value per column in bold. ``-'' denotes no LLM tokens. NL-PDDL makes no LLM calls on these variants (reported as $0$ tokens in~\citep{liu2026nlpddl}), and Fast Downward uses no LLM.}
\label{tab:closed_world}
\footnotesize
\setlength{\tabcolsep}{3pt}
\begin{tabularx}{\linewidth}{l rY rY rY rY}
\toprule
 & \multicolumn{2}{c}{Plain} & \multicolumn{2}{c}{Mystery} & \multicolumn{2}{c}{Randomized} & \multicolumn{2}{c}{Misalign.} \\
\cmidrule(lr){2-3}\cmidrule(lr){4-5}\cmidrule(lr){6-7}\cmidrule(lr){8-9}
Method & \# Tok. & SR & \# Tok. & SR & \# Tok. & SR & \# Tok. & SR \\
\midrule
GPT-4o                              & 982{,}602 & 34\% & 834{,}890 &  0\% & 835{,}974 &  0\% & 937{,}905 & 27\% \\
Gemini-2.0 Flash                    & 928{,}024 & 18\% & 834{,}060 &  1\% & 835{,}824 &  0\% & 939{,}950 & 23\% \\
LLaMA-3.1                           & 963{,}565 & 44\% & 841{,}791 &  0\% & 842{,}622 &  0\% & 950{,}490 & 42\% \\
NL-PDDL~\citep{liu2026nlpddl}       &         - & 70\% &         - & 70\% &         - & 70\% &  11{,}656 & 70\% \\
Fast Downward (upper bound)         &       - & \textbf{100}\% & - & \textbf{100}\% & - & \textbf{100}\% & - & 0\% \\
\midrule
\ourmethod\ (Qwen3-32B)             & \textbf{273{,}400} & 99.7\% & 1{,}690{,}032 & 74.0\% & 456{,}100 & 78.7\% & \textbf{341{,}400} & \textbf{100}\% \\
\ourmethod\ (gpt-oss-120b)          & 256{,}500 & \textbf{100}\% & 363{,}100 & \textbf{76.7}\% & \textbf{301{,}100} & \textbf{100}\% & 398{,}800 & 98\% \\
\bottomrule
\end{tabularx}
\end{table}

\paragraph{Token cost.}
EvoPlan's tokens-to-solve are higher than NL-PDDL's on the shared-vocabulary variants. This is structural rather than incidental: NL-PDDL is a symbolic FOL regression planner that calls the LLM only for commonsense-entailment queries (cached and reused) when the goal and domain vocabularies disagree, while EvoPlan uses the LLM as its plan proposer at every iteration with the validator providing repair feedback. The per-iteration LLM cost is the price of a few capabilities that the symbolic planner does not offer: the proposer tolerates renamed action-model vocabularies without per-variant entailment engineering, solves tasks whose optimal plans are deeper than those the regression planner solves within its runtime limit (it fails beyond depth $10$~\citep{liu2026nlpddl}), runs against any input PDDL domain $D$ without action-schema mining, and is backbone-agnostic, so quality follows the public LLM ecosystem. On problems where symbolic regression succeeds outright, EvoPlan is the more expensive choice. On problems where it does not (deep plans, surface-form renaming, vocabulary misalignment without manual entailment layers), the same budget yields a working plan within an unchanged loop. Cost grows with task difficulty rather than method complexity.

\subsection{Closed-world ALFWorld results}
\label{app:planner_alfworld}

Beyond the synthetic Blocksworld variants, we evaluate on the realistic closed-world ALFWorld domain with the canonical \texttt{alfred} PDDL action model and the $N{=}134$ \texttt{valid\_unseen} tasks of~\citet{liu2026nlpddl}. Each instance encodes the full ground-truth scene as the initial state, with existentially-typed goals. This isolates the symbolic planner from perception and exploration, which are tested separately under the open-world setting (Sec.~\ref{sec:exp_planner}). Table~\ref{tab:closed_alfworld} reports aggregate SR. The closed-world figure ($90.3\%$ for Qwen3-32B) and the open-world figure ($98.5\%$, Sec.~\ref{sec:exp_planner}) measure different protocols and are not directly comparable: the former isolates symbolic search, the latter measures the integrated explore--plan--act loop.

\begin{table}[!ht]
\centering
\caption{Closed-world ALFWorld plan-synthesis on the full $N{=}134$ \texttt{valid\_unseen} task set~\citep{liu2026nlpddl}. Fast Downward~\citep{helmert2006fastdownward} is the symbolic upper bound. ``-'': no LLM is used.}
\label{tab:closed_alfworld}
\small
\setlength{\tabcolsep}{6pt}
\begin{tabular}{lrc}
\toprule
Method & Total tokens ($N{=}134$) & SR \\
\midrule
Fast Downward (upper bound)   & -            & \textbf{100\%} \\
GPT-5.5 (LLM+VAL loop)        & 2{,}770{,}945 & \textbf{100\%} \\
\midrule
\ourmethod\ (Qwen3-32B)       & 10{,}349{,}634 & 90.3\% \\
\ourmethod\ (gpt-oss-120b)    & 17{,}071{,}316 & 78.4\% \\
\bottomrule
\end{tabular}
\vspace{-0.5em}
\end{table}

\subsection{Same-backbone planner ablation}
\label{app:planner_ablation}

We isolate the evolutionary loop's contribution by holding the backbone (Qwen3-32B), the prompt template, and the per-task budget fixed across two arms that differ only in the search loop: single-shot LLM+VAL ($k{=}1$) versus the full feedback-conditioned evolutionary loop. Closed-world Blocksworld \emph{Plain} is the discriminating benchmark because Qwen3-32B's single-shot SR there ($86.7\%$) is well below the full-EvoPlan figure ($99.7\%$), so any lift is attributable to the loop alone.

\begin{table}[!ht]
\centering
\caption{Same-backbone ablation on closed-world Blocksworld \emph{Plain} ($N{=}300$, Qwen3-32B, $10$-iteration budget). Tokens are reported per-$100$ instances to match the basis of Table~\ref{tab:closed_world}. The evolutionary loop closes a $13$\,pp SR gap over single-shot at ${\sim}1.5\times$ the token cost.}
\label{tab:planner_ablation_bw}
\small
\begin{tabular}{l c c}
\toprule
Method & Tokens-to-solve / $100$ inst.\ $\downarrow$ & SR \% $\uparrow$ \\
\midrule
Single-shot LLM+VAL ($k{=}1$, no evolution)                          & \textbf{184{,}178} & 86.7 \\
EvoPlan (feedback-conditioned evolution, $\textsc{iters}{=}10$)      & 273{,}400          & \textbf{99.7} \\
\bottomrule
\end{tabular}
\end{table}

Single-shot Qwen3-32B already beats every direct-LLM baseline and the NL-PDDL symbolic floor on Plain, confirming the backbone is strong. Adding the evolutionary loop with descriptive feedback lifts SR by $13$\,pp at roughly $1.5\times$ the token budget. The lift comes from the harder instances where the first proposal fails type-checking or omits a precondition: the validator artifact names the failed action, and the next iteration produces a targeted repair rather than another untargeted guess. Most tasks converge within $1$--$3$ iterations (App.~\ref{app:planner_blocksworld}).

\paragraph{Population search versus a single-candidate repair loop.} On Blocksworld \emph{Mystery} ($N{=}100$, Qwen3-32B, $20$-iteration budget), we compare EvoPlan with a validate-and-repair loop that repairs a single candidate with the same validator feedback and keeps no population. The repair loop solves $88/100$ instances. Two population-search runs solve $82/100$ and $74/100$. On the $67$ instances solved by all three runs, population-search plans exceed the Fast Downward optimum by $1.25$ actions on average (two-run mean) against $2.24$ for the repair loop (paired $t{=}2.64$, $p{<}0.05$), a $44\%$ reduction, and $53\%$ against $48\%$ of plans are exactly optimal. Population search therefore does not raise success rate on this variant, but it returns shorter plans.

\subsection{Fast Downward on misaligned open-world ALFWorld}
\label{app:planner_fd_openworld}

We replace the evolutionary planner with Fast Downward in the same open-world explore--plan loop on the misaligned variant ($N{=}134$), without the predicate map $\mu$. Fast Downward solves every task whose goal predicates exist in the action model ($18/18$ \texttt{look\_at\_obj\_in\_light}, $24/24$ \texttt{pick\_and\_place\_simple}, $17/17$ \texttt{pick\_two\_obj\_and\_place}) and none whose goal predicates do not ($0/31$ clean, $0/21$ cool, $0/23$ heat), $59/134 = 44.0\%$ overall. A goal predicate absent from the action model makes the problem unsolvable as written, so the failure is complete rather than gradual. With $\mu$, \ourmethod\ solves $98.5\%$ of the same tasks (Table~\ref{tab:open_world}).

\section{Constrained execution loop: runtime and integration}
\label{app:loop}

\subsection{Evolutionary planning loop pseudocode}
\label{app:loop_evolve_pseudocode}
The runtime planner maintains a database of candidate plans and their evaluator feedback. The loop uses the global $\Phi_{\mathrm{mob}}$ mined in App.~\ref{app:mining} and introduces no per-skill contracts.
\begin{lstlisting}[style=compactcode]
def evolve_plan(task, world, D, Phi_mob):
    population = seed_candidates(task, world, D)
    verified_prefix = []
    while budget_remaining():
        parent = sample(population)
        child = llm_mutate(parent, feedback=parent.feedback)
        if not pddl_syntax_ok(child):
            population.add(child, score=-inf, feedback="format error")
            continue
        val_report = validate_pddl(child, world)
        if not val_report.success:
            population.add(child, score=val_report.score, feedback=val_report.message)
            continue
        stl_report = check_phi_mob(child, world, Phi_mob)
        score = aggregate(val_report, stl_report)
        population.add(child, score=score, feedback=all_reports)
        if child.is_goal_reaching and stl_report.feasible:
            return child
    return best_safe_prefix(population)
\end{lstlisting}

\subsection{Runtime execution pseudocode}
\label{app:loop_runtime_pseudocode}
Once a candidate plan is selected, each PDDL action is dispatched action-by-action: Nav2 grounds the action into a waypoint sequence, the sequence is checked against $\Phi_{\mathrm{mob}}$, and execution proceeds only if the check succeeds. Symbolic effects come from the fixed PDDL domain $D$ supplied as input. No per-skill effect certification is performed.
\begin{lstlisting}[style=compactcode]
def execute_action(a, world, Phi_mob):
    if not symbolic_preconditions_hold(a, world):
        return emit_failure("precondition-failed")
    waypoints = nav2_plan_route(a.target, world, costmap=world.costmap)
    rho = stl_robustness(waypoints, Phi_mob)
    if rho < 0:
        return emit_failure(failure_fact_from(rho, waypoints))
    dispatch(waypoints)
    return assert_effects(a.effects)  # from PDDL domain D
\end{lstlisting}

\subsection{Failure facts}
\label{app:loop_failure_facts}
Failure facts are the interface between the continuous executor and the symbolic planner. The two emitted by the runs in this paper are \texttt{route-blocked(route2)} (Nav2 returns no admissible global path, or the clearance monitor reports a violation on the route) and \texttt{no-safe-approach(robot1,human1)} (the only admissible routes pass within the clearance floor of an observed pedestrian). The interface is extensible to additional facts (e.g.\ \texttt{low-battery}, \texttt{terrain-risk-high}, \texttt{human-location-uncertain}). These are not used by the current experiments. Emitted facts are added to the next PDDL problem instance so the planner can choose alternate actions rather than repeatedly attempting the failed one.

\subsection{Nav2 navigation interface}
\label{app:loop_nav2}
The runtime bridge converts the evolved STL-constrained plan into actual robot motion through Nav2. Each PDDL \texttt{move} action is dispatched as a navigation goal to Nav2's behaviour-tree navigator. The global planner produces a route over the static costmap and the local planner refines it into a waypoint sequence over a short horizon. The grounded $\Phi_{\mathrm{mob}}$ atoms are evaluated on this sequence before the controller executes it: a sequence that violates a clause (clearance, speed, time-to-collision) is not executed, and the planner re-plans.

\subsection{Real-robot deployment details}
\label{app:loop_real_robot}

\paragraph{Platform.} The robot is a Clearpath Jackal J100 running ROS\,2 Humble and Nav2. The office is modelled as a six-region graph with seven undirected edges and move variants for doorways and narrow passages. Perception uses YOLO-World on the robot's own tracker.

\paragraph{Planner.} The evolutionary PDDL planner of Sec.~\ref{sec:planning} is served by gpt-5-mini (low reasoning effort, diff-based evolution in OpenEvolve) in these runs, not by a locally-hosted open-weight model. VAL acts only as a gate: it truncates a plan to its executable prefix and accepts or rejects each repair. No classical planner is in the loop.

\paragraph{Monitor.} The human-avoidance replans are triggered by a geometric clearance monitor with a $1.00$\,m floor, not by the mined $\Phi_{\mathrm{mob}}$.

\paragraph{Timing ranges.} Over the six complete runs of both demonstrations: reactive tier $202$--$241$\,ms ($n{=}28$). Deliberative tier: $10.0$--$20.0$\,s ($n{=}7$). Plan swap: maximum $12.9$\,ms ($n{=}9$).

\paragraph{Perception-grounding runs.} We report two runs that completed the full survey--hold--repair--complete cycle and present them as demonstrations rather than as a success rate. Table~\ref{tab:robot_backpack_runs} compares them. In both, the detector grounded the backpack at the same region, the repaired problem gained the inspection goal for the backpack, and the robot held for $18.0$\,s while the repair was computed. The two $15$-action repaired plans differ in action order, so they are independent planner outputs, and both start at the region where the robot was parked rather than at the problem's authored start. Phase 1 never produces a VAL-valid plan, by design: the goal object has no location, so VAL truncates the plan to its executable prefix, the survey. Phase 1 ran $3$ iterations ($63$--$68$\,s wall clock). Each repair was warm-started from the deployed plan and was VAL-valid at its first iteration ($16.2$\,s of model time, $17$--$18$\,s wall clock).

\begin{table}[!ht]
\centering\small
\caption{Two complete perception-grounding runs, with times relative to each session's first event.}
\label{tab:robot_backpack_runs}
\begin{tabular}{@{}lcc@{}}
\toprule
 & Run 1 & Run 2 \\
\midrule
Survey prefix deployed            & $t{+}66.2$\,s  & $t{+}61.7$\,s \\
Object grounded, hold asserted    & $t{+}83.5$\,s  & $t{+}85.1$\,s \\
Repaired plan deployed            & $t{+}101.5$\,s & $t{+}103.1$\,s \\
Mission complete                  & $t{+}225.1$\,s & $t{+}214.3$\,s \\
Hold duration                     & $18.0$\,s      & $18.0$\,s \\
Legs / inspection dwells          & $9/8$          & $10/10$ \\
Repaired plan length              & $15$ actions   & $15$ actions \\
Distinct human tracks             & $14$           & $10$ \\
Reactive replans                  & $8$            & $8$ \\
\bottomrule
\end{tabular}
\end{table}

\paragraph{Human-avoidance runs.} A person steps into the robot's path. The clearance monitor reports a signed margin against the $1.00$\,m floor. On a violation, the reactive tier paints a $1.0$\,m lethal disc at the person's position into the global costmap and re-sends the same waypoints, and the deliberative tier parks the robot and swaps in a VAL-gated plan. We report four runs that completed the cycle (Table~\ref{tab:robot_avoid}), one of them twice after two separate encounters, and present them as demonstrations rather than as a success rate. Across the six complete runs of both demonstrations, $25$ of the $28$ costmap discs were removed automatically once the person's track expired. All runs reused one phase-1 plan and one repaired plan generated in advance, so these runs measure the runtime tiers, not the planner. The planner measurements are those of the perception-grounding runs above.

\begin{table}[!ht]
\centering\small
\caption{Human-avoidance runs on the Jackal robot. Replans counts clearance-triggered reactive replans, Viol. counts violation samples, and Min margin is the most negative clearance margin against the $1.00$\,m floor.}
\label{tab:robot_avoid}
\setlength{\tabcolsep}{4pt}
\begin{tabular}{@{}lcccccl@{}}
\toprule
Run & Window [s] & Legs & Replans & Viol. & Min margin [m] & Outcome \\
\midrule
1 & 102.9 & 6 & 4 & 44 & $-0.975$ & complete \\
2 & 505.4 & 7 & 4 & 7  & $-0.736$ & complete ($\times2$) \\
3 & 102.8 & 6 & 2 & 2  & $-0.667$ & complete \\
4 & 100.7 & 6 & 2 & 42 & $-0.939$ & complete \\
\bottomrule
\end{tabular}
\end{table}

\paragraph{Caveats.} The node used poses from the odometry frame while the region graph is in the map frame, so region labels in the event stream can drift. Timestamps and latencies are unaffected. With two complete perception-grounding runs, the results are an existence proof, not a distribution.

\subsection{Gazebo replanning missions}
\label{app:loop_gazebo_missions}

The robot localises with AMCL and drives Nav2 waypoints lowered from the PDDL plan. In the find-an-object mission the survey covers three regions. In the find-and-inspect mission it covers five, and the world holds two cones. Detections are clustered into individual objects, each object becomes a new target in the problem's objects, initial state, and goal, and each must be driven to and held in view for $5$\,s. Table~\ref{tab:gazebo_missions} reports five trials per mission.

\begin{table}[!ht]
\centering\small
\caption{Gazebo replanning missions, five trials each (mean $\pm$ std).}
\label{tab:gazebo_missions}
\begin{tabular}{@{}lcc@{}}
\toprule
 & Find-an-object & Find-and-inspect \\
\midrule
Mission complete       & $4/5$          & $3/5$ \\
Time [s]               & $227\pm15$     & $449\pm188$ \\
Deliberation [s]       & $80\pm76$      & $427\pm335$ \\
Path [m]               & $85.1\pm4.1$   & $131.8\pm56.4$ \\
Collisions             & $0.0\pm0.0$    & $0.0\pm0.0$ \\
Replans applied        & $1.2\pm0.4$    & $2.6\pm1.8$ \\
Replans abandoned      & $0.4\pm0.9$    & $1.2\pm1.6$ \\
Inspection rounds      & ---            & $2.6\pm2.3$ \\
Nav2 goal aborts       & $0.8\pm1.1$    & $3.8\pm3.9$ \\
\bottomrule
\end{tabular}
\end{table}

Every returned plan validated against the domain, yet the failing trials carry $4$--$10$ Nav2 goal aborts: the robot could not execute a correct plan from where it stood, so the remaining failure mode lies below the symbolic interface. Each recovery in the find-and-inspect mission is an LLM round trip of $50$--$80$\,s, which is why its deliberation time is close to its total time. Two confounds limit the comparison between the missions: the find-an-object mission kept a Fast Downward fallback and the find-and-inspect mission did not, and the two surveys cover different numbers of regions.

\paragraph{Human-avoidance runs.} These four runs use a geometric clearance monitor at $1.25$\,m, not the mined $\Phi_{\mathrm{mob}}$, which recorded no violations because stationary pedestrians keep time-to-collision large. On a violation, the failed route leg is retracted from the problem state and Fast Downward returns the new plan, which VAL validates. Median replan latency is $0.104$\,s (range $0.075$--$45.1$\,s, the maximum being a deadline timeout on an unsolvable problem). Each run applies two replans and covers all three regions with zero collisions. One run stopped $0.55$\,m short of its final goal.

\subsection{Crowd-density ablation in Gazebo}
\label{app:loop_crowd_sweep}

Fig.~\ref{fig:crowd_density_sweep} (Sec.~\ref{sec:exp_endtoend}) shows how the constrained execution loop responds as crowd size grows. We sweep the number of walking pedestrians in the factory scene from $0$ to $50$ ($\rho \in [0,\,0.041]\,\text{m}^{-2}$) while holding the five missions, the mined $\Phi_{\mathrm{mob}}$, and the Nav2 stack fixed. Each setting is averaged over multiple seeded runs.

At low density ($10$ pedestrians, $\rho \approx 0.008$) the loop preserves a $\sim 1$\,m minimum clearance and produces no collisions, with personal-zone exposure under $1\%$ of ticks. As density rises beyond $\rho \approx 0.016$, the closest approach drops sharply into the intimate band: the local planner has no admissible corridor that stays outside the $1.2$\,m personal zone, so the robot trades clearance margin to keep making progress. The intimate-zone occupancy in panel~(a) nonetheless remains under $2\%$ of ticks even at the densest setting, because the loop only briefly touches the intimate threshold when slipping past a pedestrian. Cumulative personal-space intrusion ($P_{\mathrm{int}}$) grows monotonically with density, reflecting the steady accumulation of small shortfalls below $1.2$\,m rather than a few catastrophic intrusions. Collisions in panel~(d) appear once density exceeds $\rho \approx 0.016$ and rise non-monotonically with population: at $50$ pedestrians the loop averages $\sim 2.5$ collisions per run, which is the operating regime where Nav2's reactive local planner cannot resolve the dense cluster within the replanning budget.

\paragraph{What the sweep shows.}
The loop degrades gracefully rather than collapsing: success-relevant metrics (clearance, intimate occupancy, collisions) move smoothly with density and do not exhibit a step-change failure. Personal space is preserved at low density ($\rho \approx 0.008$, about one person per $120\,\text{m}^2$). Beyond $\rho \approx 0.016$, the constraint is increasingly violated in a controlled way: $P_{\mathrm{int}}$ grows but the intimate-zone-occupancy ceiling holds. The non-monotonic collision count in panel~(d) reflects sensitivity to pedestrian routes at each density rather than method-level non-determinism.
\section{Extended related work}
\label{app:related_extended}

\paragraph{LLM and VLM planning.}
Early embodied LLM systems showed that language models can sequence high-level robot actions, but purely generative plans are difficult to verify. LLM+P and related approaches use formal planners as back ends, while VLMFP extends this idea to visual inputs by generating PDDL problem and domain files and iteratively aligning symbolic execution with simulated visual outcomes~\citep{hao2026simulation}. Our work shares the view that foundation models are most reliable when paired with external evaluators, but it shifts the evaluation target from only symbolic PDDL files to a hybrid PDDL plan plus a global STL constraint $\Phi_{\mathrm{mob}}$.

\paragraph{Open-world symbolic planning.}
NL-PDDL addresses incomplete knowledge and goal-action language mismatch by representing predicates in natural language and performing regression planning with LLM-based entailment~\citep{liu2026nlpddl}. This is complementary to our system. NL-PDDL expands the expressivity and accessibility of symbolic planning. Our approach focuses on mining a global mobility constraint from trajectories and monitoring it on a robot.

\paragraph{Safety-aware PDDL and verifier feedback.}
SafeGen-LLM constructs PDDL3 benchmarks with explicit safety constraints and trains LLMs using rewards derived from formal verification~\citep{fan2026safegen}. The reward categories in that work, such as format error, precondition violation, safety violation, goal not satisfied, and success, inspire our evaluator cascade. Our safety constraints are not only PDDL3 plan constraints. They are STL formulas over measured robot/environment traces.

\paragraph{Multi-robot and grounded collaboration.}
PIP-LLM combines PDDL team-level plans with integer-programming robot assignment~\citep{shi2025pipllm}. SPINE-HT grounds LLM-generated subtasks in heterogeneous robot capabilities and adapts subtasks from online feedback in unstructured environments~\citep{ravichandran2025spineht}. These systems motivate our emphasis on grounding and feedback. We focus on a single Jackal robot deployment, but the same global $\Phi_{\mathrm{mob}}$ shield and evolutionary planner can extend to multi-robot missions with added role assignment, communication, and shared-resource constraints.

\paragraph{Learning temporal logic.}
Several works learn temporal-logic structures or parameters from data. The closest project reference learns STL predicates from trajectory data using expression optimization and conformal prediction to obtain finite-sample correctness guarantees~\citep{soroka2025learning}. Our constraint-mining phase uses a similar grammar-and-evaluator view, but the learned formula is a single global mobility constraint applied uniformly across every PDDL action at runtime.

\paragraph{Semantic safety and constitutions.}
Robot constitutions and semantic safety benchmarks provide high-level rules and datasets for embodied safety~\citep{sermanet2025robotconstitutions}. Such rules are valuable for selecting or rejecting candidate instructions. Our STL constraint serves a different layer: after a task is judged appropriate, the executed robot trace must still maintain concrete measurable constraints.

\end{document}